\documentstyle[lingmacros,fullname]{article}

\title{An Alternative Conception of\\Tree-Adjoining Derivation}

\author{\begin{tabular}{c@{\qquad}c}
		 Yves Schabes & Stuart M. Shieber \\
		 Mitsubishi Electric & Division of Applied Sciences \\
		 Research Laboratories, Inc. & Harvard University \\
		 Cambridge, MA  02139& Cambridge, MA  02138\\
		{\tt shieber@das.harvard.edu} & {\tt schabes@merl.com}
        \end{tabular}}

\date{December 20, 1993}

\input{psfig-scale}

\makeatletter
\def\fnum@figure{{\bf Figure \thefigure}}
\makeatother

\newcommand{\seq}[1]{\langle #1 \rangle}
\newcommand{\set}[1]{\{ #1 \} }

\newcommand{\eitem}[7]{\langle{#1 \ra #2{} \bullet {}#3,
	\indices{#4}{#5}{#6}{#7}}\rangle}
\newcommand{\ditem}[8]{\langle{#1 \ra #2{} \bullet {}#3,
	\indices{#4}{#5}{#6}{#7}, #8}\rangle}
\newcommand{\itemoffset}{\hspace*{.5in}}

\newcommand{\unreducedra}[2]{\overline{#1 \ra #2}}
\newcommand{\unreduced}[1]{\overline{#1}}

\newcommand{\indices}[4]{#1, #2, #3, #4}

\newcommand{\good}{\makebox[.15in]{\hfill}}
\newcommand{\bad}{\makebox[.15in]{*\hfill}}
\newcommand{\unfel}{\makebox[.15in]{\#\hfill}}

\newcommand{\word}[1]{{$\langle\mbox{\it #1}\rangle$}}

\newcommand{\oneovermodb}[3]{\begin{array}[b]{@{}c@{}}
                                   #2 \\[-1.8ex]
                                   \hbox to #1{\hrulefill} \\[-.8ex]
                                   #3 \end{array}}

\newcommand{\oneover}[2]{\begin{array}{@{}c@{}} #1 \\
\hline #2  \end{array}}

\newcommand{\oneovermod}[3]{\begin{array}{@{}c@{}}
                                   #2 \\[-1.8ex]
                                   \hbox to #1{\hrulefill} \\[-.8ex]
                                   #3 \end{array}}

\newcommand{\Itree}{{\cal I}}
\newcommand{\Atree}{{\cal A}}
\newcommand{\Etree}{{\cal E}}
\newcommand{\DD}{{\cal D}}
\newcommand{\prefix}{\prec}
\newcommand{\prefixeq}{\preceq}
\newcommand{\dumarg}{\_}

\newcommand{\ra}{\rightarrow}
\newcommand{\Ra}{\Rightarrow}

\newcommand{\eqpunc}[1]{{\makebox[0pt][l]{\qquad\rm{#1}}}}
\newcommand{\conc}{\cdot}

\newcommand{\qed}{\hspace*{\fill} \mbox{$\Box$}}

\newtheorem{proposition}{Proposition}
\newtheorem{lemmanonum}{Lemma}
\newcommand{\proof}{\noindent {\bf Proof: }}

\newcommand{\qbox}[1]{\begin{center}\mbox{#1}\end{center}}

\begin{document}

\begin{titlepage}

\maketitle
\thispagestyle{empty}

\begin{abstract}
The precise formulation of derivation for tree-adjoining grammars has
important ramifications for a wide variety of uses of the formalism,
from syntactic analysis to semantic interpretation and statistical
language modeling.  We argue that the definition of tree-adjoining
derivation must be reformulated in order to manifest the proper
linguistic dependencies in derivations.  The particular proposal is
both precisely characterizable through a definition of TAG derivations
as equivalence classes of ordered derivation trees, and
computationally operational, by virtue of a compilation to linear
indexed grammars together with an efficient algorithm for recognition
and parsing according to the compiled grammar.
\end{abstract}

\vfill

{\noindent\small This paper is to appear in {\em Computational
Linguistics}, volume 20, number 1, and is available from the Center for
Research in Computing Technology, Division of Applied Sciences,
Harvard University as Technical Report TR-08-92 and through the
Computation and Language e-print archive as cmp-lg/9404001.}

\end{titlepage}

\tableofcontents \newpage

\section{Introduction}

In a context-free grammar, the derivation of a string in the rewriting
sense can be captured in a single canonical tree structure that
abstracts all possible derivation orders.  As it turns out, this {\em
derivation tree\/} also corresponds exactly to the hierarchical structure
that the derivation imposes on the string, the {\em derived tree\/}
structure of the string.  The formalism of tree-adjoining grammars
(TAG), on the other hand, decouples these two notions of derivation tree
and derived tree.  Intuitively, the derivation tree is a more finely
grained structure than the derived tree, and as such can serve as a
substrate on which to pursue further analysis of the string.  This
intuitive possibility is made manifest in several ways.  Fine-grained
syntactic analysis can be pursued by imposing on the derivation tree
further combinatorial constraints, for instance, selective adjoining
constraints or equational constraints over feature structures.
Statistical analysis can be explored through the specification of
derivational probabilities as formalized in stochastic tree-adjoining
grammars.  Semantic analysis can be overlaid through the synchronous
derivations of two TAGs.

All of these methods rely on the derivation tree as the source of the
important primitive relationships among trees.  The decoupling of
derivation trees from derived trees thus makes possible a more flexible
ability to pursue these types of analyses.  At the same time, the exact
definition of derivation becomes of paramount importance.  In this
paper, we argue that previous definitions of tree-adjoining derivation
have not taken full advantage of this decoupling, and are not as
appropriate as they might be for the kind of further analysis that
tree-adjoining analyses could make possible.  In particular, the
standard definition of derivation, due to Vijay-Shanker
\shortcite{v87}, requires that auxiliary trees be adjoined at
distinct nodes in elementary trees.  However, in certain cases,
especially cases characterized as linguistic modification, it is more
appropriate to allow multiple adjunctions at a single node.

In this paper, we propose a redefinition of TAG derivation along these
lines, whereby multiple auxiliary trees of modification can be
adjoined at a single node, whereas only a single auxiliary tree of
predication can.  The redefinition constitutes a new definition of
derivation for TAG that we will refer to as {\em extended derivation}.
In order for such a redefinition to be serviceable, however, it is
necessary that it be both precise and operational.  In service of the
former, we provide a formal definition of extended derivation using a
new approach to representing derivations as equivalence classes of
ordered derivation trees.  With respect to the latter, we provide a
method of compilation of TAGs into corresponding linear indexed
grammars (LIG), which makes the derivation structure explicit, and
show how the generated LIG can drive a parsing algorithm that
recovers, either implicitly or explicitly, the extended derivations of
the string.

The paper is organized as follows.  First, we review Vijay-Shanker's
standard definition of TAG derivation, and introduce the motivation
for extended derivations.  Then, we present the extended notion of
derivation and its formal definition.  The original compilation of
TAGs to LIGs provided by Vijay-Shanker and Weir and our variant for
extended derivations are both described.  Finally, we discuss a
parsing algorithm for TAG that operates by a variant of Earley parsing
on the corresponding LIG.  The set of extended derivations can
subsequently be recovered from the set of Earley items generated by
the algorithm.  The resultant algorithm is further modified so as to
build an explicit derivation tree incrementally as parsing proceeds;
this modification, which is a novel result in its own right, allows
the parsing algorithm to be used by systems that require incremental
processing with respect to tree-adjoining grammars.

\section{The Standard Definition of Derivation}

\begin{figure}
{\qbox{\psfig{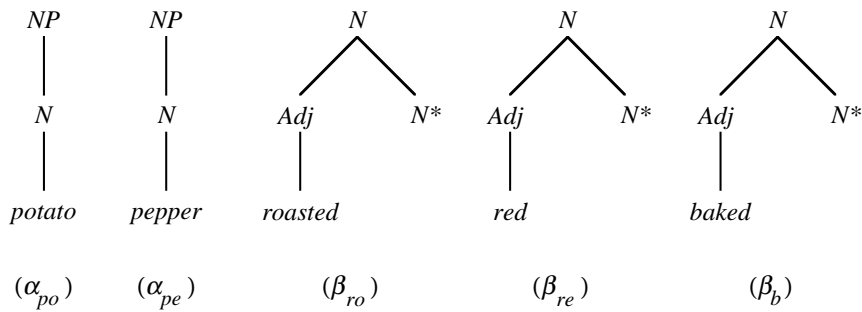}}}
\caption{A sample tree-adjoining grammar}
\label{fig:baked}
\end{figure}

\begin{figure}
{\qbox{\psfig{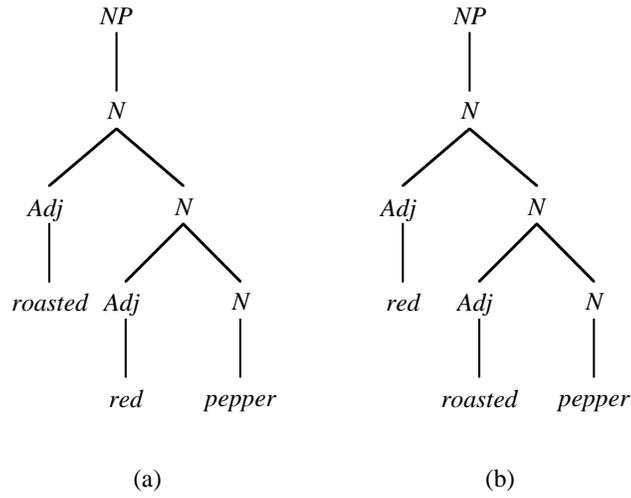}}}
\caption{Two trees derived by the grammar of Figure~\protect\ref{fig:baked}}
\label{fig:derived}
\end{figure}

To exemplify the distinction between standard and extended
derivations, we exhibit the TAG of
Figure~\ref{fig:baked}.\footnote{Here and elsewhere, we conventionally
use the Greek letter $\alpha$ and its subscripted and primed variants
for initial trees, $\beta$ and its variants for auxiliary trees, and
$\gamma$ and its variants for elementary trees in general.  The foot
node of an auxiliary tree is marked with an asterisk (`*').} This
grammar derives some simple noun phrases such as ``roasted red
pepper'' and ``baked red potato''.  The former, for instance, is
associated with the derived tree in Figure~\ref{fig:derived}(a).  The
tree can be viewed as being derived in two ways\footnote{We ignore
here the possibility of another dependent derivation wherein
adjunction occurs at the foot node of an auxiliary tree.  Because this
introduces yet another systematic ambiguity, it is typically
disallowed by stipulation in the literature on linguistic analyses
using TAGs.}

\begin{figure}
{\qbox{\psfig{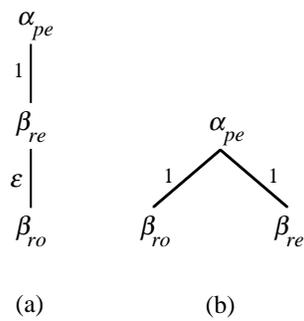}}}
\caption{Derivation trees for the derived tree of
Figure~\protect\ref{fig:derived}(a) according to the grammar of
Figure~\protect\ref{fig:baked}}
\label{fig:derivations}
\end{figure}

\begin{description}

\item[Dependent:] The auxiliary tree $\beta_{ro}$ is adjoined at the root
node (address $\epsilon$)\footnote{The address of a node in a tree is
taken to be its Gorn number, that sequence of integers specifying
which branches to traverse in order starting from the root of the tree
to reach the node.  The address of the root of the tree is therefore
the empty sequence, notated $\epsilon$.  See the appendix for a more
complete discussion of notation.} of $\beta_{re}$. The resultant tree
is adjoined at the $N$ node (address $1$) of initial tree
$\alpha_{pe}$.  This derivation is depicted as the derivation tree in
Figure~\ref{fig:derivations}(a).

\item[Independent:] The auxiliary trees $\beta_{ro}$ and  $\beta_{re}$ are
adjoined at the $N$ node (address $1$) of the initial tree
$\alpha_{pe}$.  This derivation is depicted as the derivation tree in
Figure~\ref{fig:derivations}(b).

\end{description}

\noindent In the {\em independent\/} derivation, two trees are separately
adjoined at one and the same node in the initial tree.  In the {\em
dependent\/} derivation, on the other hand, one auxiliary tree is adjoined
to the other, the latter only being adjoined to the initial tree.  We
will use this informal terminology uniformly in the sequel to
distinguish the two general topologies of derivation trees.

The {\em standard\/} definition of derivation, as codified by
Vijay-Shanker, restricts derivations so that {\em two adjunctions
cannot occur at the same node in the same elementary tree}.  The
dependent notion of derivation (Figure~\ref{fig:derivations}(a)) is
therefore the only sanctioned derivation for the desired tree in
Figure~\ref{fig:derived}(a); the independent derivation
(Figure~\ref{fig:derivations}(b)) is disallowed.  Vijay-Shanker's
definition is appropriate because for any independent derivation,
there is a dependent derivation of the same derived tree.  This can be
easily seen in that any adjunction of $\beta_2$ at a node at which an
adjunction of $\beta_1$ occurs could instead be replaced by an
adjunction of $\beta_2$ at the root of $\beta_1$.

The advantage of this {standard} definition of derivation is that a
derivation tree in this normal form unambiguously specifies a derived
tree.  The independent derivation tree on the other hand is ambiguous as
to the derived tree it specifies in that a notion of precedence of the
adjunctions at the same node is unspecified, but crucial to the derived
tree specified.  This follows from the fact that the independent
derivation tree is symmetric with respect to the roles of the two
auxiliary trees (by inspection), whereas the derived tree is not.  By
symmetry, therefore, it must be the case that the same independent
derivation tree specifies the alternative derived tree in
Figure~\ref{fig:derived}(b).

\section[Motivation for an Extended Definition of
Derivation]{Motivation for an Extended Definition of Derivation}

In the absence of some further interpretation of the derivation tree
nothing hinges on the choice of derivation definition, so that the
standard definition disallowing independent derivations is as
reasonable as any other.  However, tree-adjoining grammars are almost
universally extended with augmentations that make the issue apposite.
We discuss three such variations here, all of which argue for the use
of independent derivations under certain circumstances.\footnote{The
formulation of derivation for tree-adjoining grammars is also of
significance for other grammatical formalisms based on weaker forms of
adjunction such as lexicalized context-free grammar \cite{sw93-acl}
and its stochastic extension \cite{sw-iwpt93}, though we do not
discuss these arguments here.}

\subsection{Adding Adjoining Constraints}

Already in very early work on tree-adjoining grammars \cite{jlt75}
constraints were allowed to be specified as to whether a particular
auxiliary tree may or may not be adjoined at a particular node in a
particular tree.  The idea is formulated in its modern variant as {\em
selective-adjoining constraints\/} \cite{vj85}.  As an application of
this capability, we consider the traditional grammatical view that
directional adjuncts can be used only with certain verbs.\footnote{For
instance, Quirk et al.~\shortcite[page 517]{quirk-greenbaum} remark
that ``direction adjuncts of both goal and source can normally be used
only with verbs of motion''.  Although the restriction is undoubtedly
a semantic one, we will examine the modeling of it in a TAG deriving
syntactic trees for two reasons.  First, the problematic nature of
independent derivation is more easily seen in this way.  Second, much
of the intuition behind TAG analyses is based on a tight relationship
between syntactic and semantic structure.  Thus, whatever scheme for
semantics is to be used with TAGs will require appropriate derivations
to model these data.  For example, an analysis of this phenomenon by
adjoining constraints on the semantic half of a synchronous TAG would
be subject to the identical argument.  See
Section~\ref{sec:add-semantics}.} This would account for the felicity
distinctions between the following sentences:
\eenumsentence{
\item \label{eg:walk1} \good Brockway walked his Labrador
towards the yacht club.
\item \label{eg:resemble1} \unfel Brockway resembled his Labrador
towards the yacht club.
}

This could be modeled by disallowing through selective adjoining
constraints the adjunction of the elementary tree corresponding to a
{\em towards\/} adverbial at the VP node of the elementary tree corresponding
to the verb {\em resembles}.\footnote{Whether the adjunction occurs at
the VP node or the S node is immaterial to the argument.} However, the
restriction applies even with intervening (and otherwise acceptable)
adverbials.
\eenumsentence{
\item \good Brockway walked his Labrador yesterday.
\item \label{eg:walk2} \good Brockway walked his Labrador
yesterday
towards the yacht club.
}
\eenumsentence{
\item \good Brockway resembled his Labrador yesterday.
\item \label{eg:resemble2} \unfel Brockway resembled his Labrador
yesterday towards the yacht club.
}
Under the standard definition of derivation, there is no direct
adjunction in the latter sentence of the {\em towards\/} tree into the {\em
resembles\/} tree.  Rather, it is dependently adjoined at the root of
the elementary tree that heads the adverbial {\em yesterday}, the
latter directly adjoining into the main verb tree.  To restrict both
of the ill-formed sentences, then, a restriction must be placed not
only on adjoining the goal adverbial in a {\em resembles\/} context, but
also in the {\em yesterday\/} adverbial context.  But this constraint is
too strong, as it disallows sentence~(\ref{eg:walk2}) above as
well.

The problem is that the standard derivation does not correctly reflect
the syntactic relation between the adverbial modifier and the phrase
it modifies when there are multiple modifications in a single clause.
In such a case, each of the adverbials independently modifies the
verb, and this should be reflected in their independent adjunction at
the same point.  But this is specifically disallowed in a standard
derivation.

Another example along the same lines follows from the requirement that
tense as manifested in a verb group be consistent with temporal
adjuncts.  For instance, consider the following examples:
\eenumsentence{
\item \good  Brockway walked    his Labrador yesterday.
\item \unfel Brockway will walk his Labrador yesterday.
}
\eenumsentence{
\item \unfel Brockway walked    his Labrador tomorrow.
\item \good  Brockway will walk his Labrador tomorrow.
}
Again, the relationship is independent of other intervening
adjuncts.
\eenumsentence{
\item \good  Brockway walked    his Labrador towards the yacht
				club yesterday.
\item \unfel Brockway will walk his Labrador towards the yacht
				club yesterday.
}
\eenumsentence{
\item \unfel Brockway walked    his Labrador towards the yacht
				club tomorrow.
\item \good  Brockway will walk his Labrador towards the yacht
				club tomorrow.
}

It is important to note that these arguments apply specifically to
auxiliary trees that correspond to a modification relationship.
Auxiliary trees are used in TAG typically for predication relations as
well,\footnote{We use the term `predication' in its logical sense,
that is, for auxiliary trees that serve as logical predicates over the
trees into which they adjoin, in contrast to the term's linguistic
sub-sense in which the argument of the predicate is a linguistic
subject.} as in the case of raising and sentential complement
constructions.\footnote{The distinction between predicative and
modifier trees has been proposed previously for purely linguistic
reasons by Kroch \shortcite{kroch89}, who refers to them as complement
and athematic trees, respectively.  The arguments presented here can
be seen as providing further evidence for differentiating the two
kinds of auxiliary trees.  A precursor to this idea can perhaps be
seen in the distinction between repeatable and nonrepeatable
adjunction in the formalism of string adjunct grammars, a precursor of
TAGs \cite[pages 253--254]{joshi-sag2}.} Consider the following
sentences.  (The brackets mark the leaves of the pertinent trees to be
combined by adjunction in the assumed analysis.)
\eenumsentence{
\item \good Brockway assumed that Harrison wanted to walk his
Labrador.
\item \good [Brockway assumed that] [Harrison wanted] [to walk
his Labrador]
}
\eenumsentence{
\item \good Brockway wanted to try to walk his Labrador.
\item \good [Brockway wanted] [to try] [to walk his Labrador]
}
\eenumsentence{
\item \label{eg:pred-bad} \bad Harrison wanted Brockway tried to
walk his Labrador.
\item \bad [Harrison wanted] [Brockway tried] [to
walk his Labrador]
}
\eenumsentence{
\item \label{eg:pred-good} \good Harrison wanted to assume
that Brockway walked his Labrador.
\item  \good [Harrison wanted] [to assume that] [Brockway walked his
Labrador]
}
Assume (following, for instance, the analysis of Kroch and Joshi
\shortcite{kj85}) that the trees associated with the various forms of the
verbs {\em try}, {\em want}, and {\em assume\/} all take sentential
complements, certain of which are tensed with overt subjects and
others untensed with empty subjects.  The auxiliary trees for these
verbs specify by adjoining constraints which type of sentential
complement they take: {\em assume\/} requires tensed complements,
{\em want\/} and {\em try\/} untensed.  Under this analysis the auxiliary
trees must not be allowed to independently adjoin at the same node.
For instance, if trees corresponding to ``Harrison wanted'' and
``Brockway tried'' (which both require untensed complements) were both
adjoined at the root of the tree for ``to walk his Labrador'', the
selective adjoining constraints would be satisfied, yet the generated
sentence~(\ref{eg:pred-bad}) is ungrammatical.  Conversely, under
independent adjunction, the sentence~(\ref{eg:pred-good}) would be
deemed ungrammatical, although it is in fact grammatical.  Thus, the
case of predicative trees is entirely unlike that of modifier trees.
Here, the standard notion of derivation is exactly what is needed as
far as interpretation of adjoining constraints is concerned.

An alternative would be to modify the way in which adjoining
constraints are updated upon adjunction.  If after adjoining a
modifier tree at a node, the adjoining constraints of the original
node, rather than those of the root and foot of the modifier tree, are
manifest in the corresponding nodes in the derived tree, the adjoining
constraints would propagate appropriately to handle the examples
above.  This alternative leads, however, to a formalism for which
derivation trees are no longer context-free, with concomitant
difficulties in designing parsing algorithms.  Instead, the
extended definition of derivation effectively allows use of a
Kleene-* in the ``grammar'' of derivation trees.

Adjoining constraints can also be implemented using feature structure
equations \cite{vj88}.  It is possible that judicious use of such
techniques might prevent the particular problems noted here.  Such an
encoding of a solution requires consideration of constraints that pass
among many trees just to limit the cooccurrence of a pair of trees.
However, it more closely follows the spirit of TAGs to state such
intuitively local limitations locally.

In summary, the interpretation of adjoining constraints in TAG is
sensitive to the particular notion of derivation that is used.
Therefore, it can be used as a litmus test for an appropriate
definition of derivation.  As such, it argues for a nonstandard,
independent, notion of derivation for modifier auxiliary trees and a
standard, dependent, notion for predicative trees.

\subsection{Adding Statistical Parameters}

In a similar vein, the statistical parameters of a stochastic
lexicalized TAG (SLTAG) \cite{resnik91,schabes-stochastic} specify the
probability of adjunction of a given auxiliary tree at a specific node
in another tree.  This specification may again be interpreted with
regard to differing derivations, obviously with differing impact on
the resulting probabilities assigned to derivation trees.  (In the
extreme case, a constraint prohibiting adjoining corresponds to a zero
probability in an SLTAG.  The relation to the argument in the previous
section follows thereby.)  Consider a case in which linguistic
modification of noun phrases by adjectives is modeled by adjunction of
a modifying tree.  Under the standard definition of derivation,
multiple modifications of a single NP would lead to dependent
adjunctions in which a first modifier adjoins at the root of a second.
As an example, we consider again the grammar given in
Figure~\ref{fig:baked}, that admits of derivations for the strings
``baked red potato'' and ``baked red pepper''.  Specifying adjunction
probabilities on standard derivations, the distinction between the
overall probabilities for these two strings depends solely on the
adjunction probabilities of $\beta_{re}$ (the tree for {\em red}) into
$\alpha_{po}$ and $\alpha_{pe}$ (those for {\em potato\/} and {\em
pepper}, respectively), as the tree $\beta_b$ for the word {\em baked\/}
is adjoined in both cases at the root of $\beta_{re}$ in both standard
derivations.  In the extended derivations, on the other hand, both
modifying trees are adjoined independently into the noun trees.  Thus,
the overall probabilities are determined as well by the probabilities
of adjunction of the trees for {\em baked\/} into the nominal trees.  It
seems intuitively plausible that the most important relationships to
characterize statistically are those between modifier and modified,
rather than between two
modifiers.\footnote{\label{fn:experiment}Intuition is an appropriate
guide in the design of the SLTAG framework, as the idea is to set up a
linguistically plausible infrastructure on top of which a
lexically-based statistical model can be built.  In addition,
suggestive (though certainly not conclusive) evidence along these
lines can be gleaned from corpora analyses.  For instance, in a simple
experiment in which medium frequency triples of exactly the discussed
form ``\word{adjective} \word{adjective} \word{noun}'' were examined,
the mean mutual information between the first adjective and the noun
was found to be larger than that between the two adjectives.  The
statistical assumptions behind this particular experiment do not allow
very robust conclusions to be drawn, and more work is needed along
these lines.} In the case at hand, the fact that one typically refers
to the process of cooking potatoes as ``baking'', whereas the
appropriate term for the corresponding cooking process applied to
peppers is ``roasting'', would be more determining of the expected
overall probabilities.

Note again that the distinction between modifier and predicative trees
is important.  The standard definition of derivation is entirely
appropriate for adjunction probabilities for predicative trees, but not
for modifier trees.

\subsection{Adding Semantics}
\label{sec:add-semantics}

Finally, the formation of synchronous TAGs has been proposed to allow
use of TAGs in semantic interpretation, natural language generation,
and machine translation.  In previous work \cite{shieber-schabes}, the
definition of synchronous TAG derivation is given in a manner that
requires multiple adjunctions at a single node.  The need for such
derivations follows from the fact that synchronous derivations are
intended to model semantic relationships.  In cases of multiple
adjunction of modifier trees at a single node, the appropriate
semantic relationships comprise separate modifications rather than
cascaded ones, and this is reflected in the definition of synchronous
TAG derivation.\footnote{The importance of the distinction between
predicative and modifier trees with respect to how derivations are
defined was not appreciated in the earlier work; derivations were
taken to be of the independent variety in all cases.  In future work,
we plan to remedy this flaw.} Because of this, a parser for
synchronous TAGs must recover, at least implicitly, the extended
derivations of TAG derived trees.  Shieber
\shortcite{shieber-restricting} provides a more complete discussion of the
relationship between synchronous TAGs and the extended definition
of derivation with special emphasis on the ramifications for formal
expressivity.

Note that the independence of the adjunction of modifiers in the
syntax does not imply that semantically there is no precedence or
scoping relation between them.  As exemplified in
Figure~\ref{fig:ext-deriv}, the derived tree generated by multiple
independent adjunctions at a single node still manifests nesting
relationships among the adjoined trees.  This fact may be used to
advantage in the semantic half of a synchronous tree-adjoining grammar
to specify the semantic distinction between, for example, the
following two sentences:\footnote{We are indebted to an anonymous
reviewer of an earlier version of this paper for raising this issue
crisply through examples similar to those given here.}
\eenumsentence{
\item \good Brockway ran over his polo mallet twice intentionally.
\item \good Brockway ran over his polo mallet intentionally twice.
}
We hope to address this issue in greater detail in future
work on synchronous tree-adjoining grammars.

\subsection{Desired Properties of Extended Derivations}

We have presented several arguments that the standard notion of
derivation does not allow for an appropriate specification of
dependencies to be captured.  An extended notion of derivation is needed
that
\begin{enumerate}
\item Differentiates predicative and modifier auxiliary trees;
\item Requires dependent derivations for predicative trees;
\item Allows independent derivations for modifier trees; and
\item Unambiguously and nonredundantly specifies a derived tree.
\end{enumerate}
\label{sec:other-considerations}
Furthermore, following from considerations of the role of modifier
trees in a grammar as essentially optional and freely applicable
elements, we would like the following criterion to hold of extended
derivations:
\begin{enumerate}
\item[5.] If a node can be modified at all, it can be modified any
number of times, including zero times.
\end{enumerate} \label{sec:criteria}

Recall that a derivation tree (as traditionally conceived) is a tree
with unordered arcs where each node is labeled by an elementary tree
of a TAG and each arc is labeled by a tree address specifying a node
in the parent tree.  In a standard derivation tree no two sibling arcs
can be labeled with the same address.  In an extended derivation tree,
however, the condition is relaxed: No two sibling arcs {\em to
predicative trees\/} can be labeled with the same address.  Thus, for
any given address there can be at most one predicative tree and
several modifier trees adjoined at that node.  As we have seen, this
relaxed definition violates the fourth desideratum above; for
instance, the derivation tree in Figure~\ref{fig:derivations}(b)
ambiguously specifies both derived trees in Figure~\ref{fig:derived}.
In the next section, we provide a formal definition of extended
derivations that satisfies all of the criteria above.

\section{Formal Definition of Extended Derivations}

In this section we introduce a new framework for describing TAG
derivation trees that allows for a natural expression of both standard
and extended derivations, and makes available even more fine-grained
restrictions on derivation trees.  First, we define ordered derivation
trees and show that they unambiguously but redundantly specify
derivations.\footnote{Historical precedent for independent derivation
and the associated ordered derivation trees can be found in the
derivation trees postulated for string adjunct grammars \cite[pages
99--100]{joshi-sag1}.  In this system, siblings in derivation trees
are viewed as totally, not partially, ordered.  The systematic
ambiguity introduced thereby is eliminated by stipulating that the
sibling order be consistent with an arbitrary ordering on adjunction
sites.} We characterize the redundant trees as those related by a
sibling swapping operation.  Derivation trees proper are then taken to
be the equivalence classes of ordered derivation trees where the
equivalence relation is generated by the sibling swapping.  By
limiting the underlying set of ordered derivation trees in various
ways, Vijay-Shanker's definition of derivation tree, a precise form of
the extended definition, and many other definitions of derivation can
be characterized in this way.

\subsection{Ordered Derivation Trees}

\label{sec:odt}

Ordered derivation trees, like the traditional derivation trees, are
trees with nodes labeled by elementary trees where each arc
is labeled with an address in the tree for the parent node of the arc.
However, the arcs are taken to be ordered with respect to each other.

An ordered derivation tree is well-formed if for each of its arcs,
linking parent node labeled $\gamma$ to child node labeled $\gamma'$
and itself labeled with address $t$, the tree $\gamma'$ is an
auxiliary tree that can be adjoined at the node $t$ in the tree
$\gamma$.  (Alternatively, if substitution is allowed, $\gamma'$ may
be an initial tree that can be substituted at the node $t$ in
$\gamma$.  Later definitions ignore this possibility, but are easily
generalized.)

We define the function $\DD$ from ordered derivation trees to the
derived trees they specify, according to the following recursive
definition:
\[
\DD(D) = \left\{
\begin{array}{l}
\makebox[0pt][l]{$\gamma$}
\mbox{\qquad if $D$ is a trivial tree of one node labeled with the
elementary tree $\gamma$}\\[1ex]
\gamma[\DD(D_1)/t_1, \DD(D_2)/t_2, \ldots, \DD(D_k)/t_k] \\
\mbox{\qquad if\begin{tabular}[t]{l}
		$D$ is a tree with root node labeled with the
		elementary tree $\gamma$\\
		and with $k$ child subtrees $D_1, \ldots, D_k$\\
		whose arcs are labeled with addresses $t_1, \ldots, t_k$.
		\end{tabular}}
\end{array}
\right.
\]
Here $\gamma[A_1/t_1, \ldots, A_k/t_k]$ specifies the simultaneous
adjunction of trees $A_1$ through $A_k$ at $t_1$ through $t_k$,
respectively, in $\gamma$.  It is defined as the iterative adjunction of
the $A_i$ in order at their respective addresses, with appropriate
updating of the tree addresses of any later adjunction to reflect the
effect of earlier adjunctions that occur at addresses dominating the
address of the later adjunction.

\subsection{Derivation Trees}

\begin{figure}
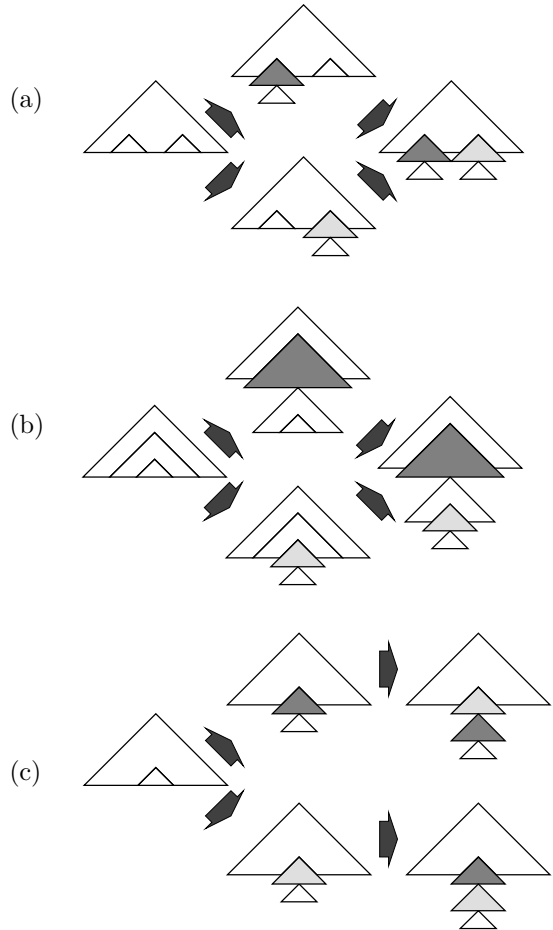

\begin{center}
\begin{tabular}{ll}
(a) &
\raisebox{-.75in}{\mbox{\psfig{figure=./graph-proof-horiz.ps,scale=50}}}\\[.95in]
(b) &
\raisebox{-.75in}{\mbox{\psfig{figure=./graph-proof-vert.ps,scale=50}}}\\[.95in]
(c) &
\raisebox{-.75in}{\mbox{\psfig{figure=./graph-proof-same.ps,scale=50}}}\\[.85in]
\end{tabular}
\end{center}
\caption{A graphical proof of the irrelevance of adjacent sibling
swapping.  These diagrams show the effect of performing two adjunctions
(of auxiliary trees depicted one as dark-shaded and one light-shaded),
presumed to be specified by adjacent siblings in an ordered derivation
tree.  The adjunctions are to occur at two addresses (referred to in
this caption as $t$ and $t'$, respectively).  The two addresses must
be such that either (a) they are distinct but neither dominates the
other, (b) $t$ dominates $t'$ (or vice versa), or (c) they are
identical.  In case (a) the diagram shows that either order of
adjunction yields the same derived tree.  Adjunction at $t$ and then
$t'$ corresponds to the upper arrows, adjunction at $t'$ and then $t$
the lower arrows.  Similarly, in case (b), adjunction at $t$ followed
by adjunction at an appropriately updated $t'$ yields the same result
as adjunction first at $t'$ and then at $t$.  Clearly, adjunctions
occurring before these two or after do not affect the
interchangeability.  Thus, if two adjacent siblings in a derivation
tree specify adjunctions at distinct addresses $t$ and $t'$, the
adjunctions can occur in either order.  Diagram (c) demonstrates that
this is not the case when $t$ and $t'$ are the same.}
\label{fig:graph-proof}
\end{figure}

It is easy to see that the derived tree specified by a given ordered
derivation tree is unchanged if adjacent siblings whose arcs are
labeled with different tree addresses are swapped.  (This is not true
of adjacent siblings whose arcs are labeled with the same address.)
That is, if $t \not = t'$ then $\gamma[ \ldots, A/t, B/t', \ldots] = \gamma[
\ldots, B/t', A/t, \ldots]$.  A graphical ``proof'' of this intuitive
fact is given in Figure~\ref{fig:graph-proof}.  A formal proof,
although tedious and unenlightening, is possible as well.  We
provide it in an appendix, primarily because the definitional aspects
of the TAG formulation may be of some interest.

This fact about the swapping of adjacent siblings shows that ordered
derivation trees possess an inherent redundancy.  The order of
adjacent sibling subtrees labeled with different tree addresses is
immaterial.  Consequently, we can define true {\em derivation trees\/}
to be the equivalence classes of the base set of ordered derivation
trees under the equivalence relation generated by the sibling subtree
swapping operation above.  This is a well-formed definition by virtue
of the proposition argued informally above.

This definition generalizes the traditional definition in not
restricting the tree address labels in any way.  It therefore
satisfies criterion (3) of Section~\ref{sec:criteria}.  Furthermore,
by virtue of the explicit quotient with respect to sibling swapping, a
derivation tree under this definition unambiguously and nonredundantly
specifies a derived tree (criterion 4).  It does not, however,
differentiate predicative from modifier trees (criterion (1)), nor can
it therefore mandate dependent derivations for predicative trees
(criterion (2)).

This general approach can, however, be specialized to correspond to
several previous definitions of derivation tree.  For instance, if we
further restrict the base set of ordered derivation trees so that no
two siblings are labeled with the same tree address, then the
equivalence relation over these ordered derivation trees allows for
full reordering of all siblings.  Clearly, these equivalence classes
are isomorphic to the unordered trees, and we have reconstructed
Vijay-Shanker's standard definition of derivation tree.

If we instead restrict ordered derivation trees so that no two
siblings corresponding to predicative trees are labeled with the same
tree address, then we have reconstructed a version of the extended
definition argued for in this paper.  Under this restriction, criteria
(1) and (2) are satisfied, while maintaining (3) and (4).

By careful selection of other constraints on the base set, other
linguistic restrictions might be imposed on derivation trees, still
using the same definition of derivation trees as equivalence classes
over ordered derivation trees.  In the next section, we show that the
definition of the previous paragraph should be further restricted to
disallow the reordering of predicative and modifier trees.  We also
describe other potential linguistic applications of the ability to
finely control the notion of derivation through the use of ordered
derivation trees.

\subsection{Further Restrictions on Extended Derivations}

The extended definition of derivation tree given in the previous
section effectively specifies the output derived tree by adding a
partial ordering on sibling arcs that correspond to modifier trees
adjoined at the same address.  All other arcs are effectively
unordered (in the sense that all relative orderings of them exist in
the equivalence class).

Assume that in a given tree $\gamma$ at a particular address $t$, the $k$
modifier trees	$\mu_1, \ldots, \mu_k$ are directly adjoined in that
order.  Associated with the subtrees rooted at the $k$ elementary
auxiliary trees in this derivation are $k$ derived auxiliary trees
($A_1, \ldots, A_k$, respectively).  The derived tree
specified by this derivation tree, according to the definition of
$\DD$ given above, would have the derived tree $A_1$ directly below
$A_2$ and so forth, with $A_k$ at the top.  Now suppose that in
addition, a predicative tree $\pi$ is also adjoined at address $t$.  It
must be ordered with respect to the $\mu_i$ in the derivation tree, and
its relative order determines where in the bottom to top order in the
derived tree the tree $A_\pi$ associated with the subderivation rooted
at $\pi$ goes.

The question that we raise here is whether all $k+1$ possible
placements of the tree $\pi$ relative to the $\mu_i$ are
linguistically reasonable.  We might allow all $k+1$ orderings (as in
the definition of the previous section), or we might restrict them by
requiring, say, that the predicative tree always be adjoined before,
or perhaps after, any modifier trees at a given address.  We emphasize
that this is a linguistic question, in the sense that the definition
of extended derivation is well-formed whatever decision is made on
this question.

\begin{figure}
{\qbox{\psfig{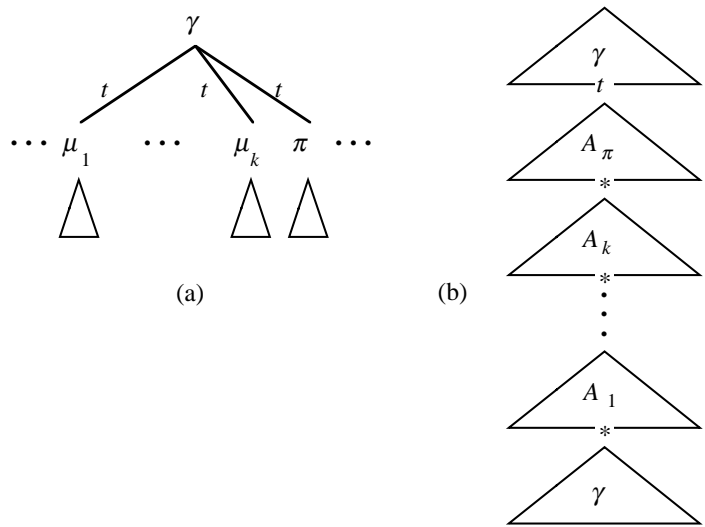}}}
\caption{Schematic extended derivation tree and associated derived
tree.  In a derived tree, the predicative tree adjoined at an address
$t$ is required to follow all modifier trees adjoined at the same
address, as in (a).  The derived tree therefore appears as depicted in
(b) with the predicative tree outermost.}
\label{fig:ext-deriv}
\end{figure}

Henceforth, we will assume that predicative trees are always adjoined
after any modifier trees at the same address, so that they appear
above the modifier trees in the derived tree.  We call this
``outermost predication'' \label{sec:inner-outer} because a
predicative tree appears wrapped around the outside of the modifier
trees adjoined at the same address.  (See Figure~\ref{fig:ext-deriv}.)
If we were to mandate innermost predication, in which a predicative
tree is always adjoined before the modifier trees at the same address,
the predicative tree would appear within all of the modifier trees,
innermost in the derived tree.

Linguistically, the outermost method specifies that if both a
predicative tree and a modifier tree are adjoined at a single node,
then the predicative tree attaches higher than the modifier tree; in
terms of the derived tree, it is as if the predicative tree were
adjoined at the root of the modifier tree.  This accords with the
semantic intuition that in such a case (for English at least), the
modifier is modifying the original tree, not the predicative one.
(The alternate ``reading'', in which the modifier modifies the
predicative tree, is still obtainable under an outermost-predication
standard by having the modifier auxiliary tree adjoin dependently at
the root node of the predicative tree.)  In contrast, the
innermost-predication method specifies that the modifier tree attaches
higher, as if the modifier tree adjoined at the root of the
predicative tree and was therefore modifying the predicative tree,
contra semantic intuitions.

For this reason, we specify that outermost predication is mandated.
This is easily done by further limiting the base set of ordered
derivation trees to those in which predicative trees are ordered after
modifier tree siblings.

(From a technical standpoint, by the way, the outermost-predication
method has the advantage that it requires no changes to the parsing
rules to be presented later, but only a single addition.  The
innermost-predication method induces some subtle interactions between
the original parsing rules and the additional one, necessitating a
much more complicated set of modifications to the original algorithm.
In fact, the complexities in generating such an algorithm constituted
the precipitating factor that led us to revise our original,
innermost-predication, attempt at redefining tree-adjoining
derivation.  The linguistic argument, although commanding, became
clear to us only later.)

Another possibility, which we mention but do not pursue here, is to
allow for language-particular precedence constraints to restrict the
possible orderings of derivation-tree siblings, in a manner similar to
the linear precedence constraints of ID/LP format \cite{gkps85} but at
the level of derivation trees.  These might be interpreted as hard
constraints or soft orderings depending on the application.  This more
fine-grained approach to the issue of ordering has several
applications.  Soft orderings might be used to account for ordering
preferences among modifiers, such as the default ordering of English
adjectives that accounts for the typical preference for ``a large red
ball'' over ``?~a red large ball'' and the typical ordering of
temporal before spatial adverbial phrases in German.

Similarly, hard constraints might allow for the handling of an
apparent counterexample to the outermost-predication
rule.\footnote{Other solutions are possible that do not require
extended derivations or linear precedence constraints.  For
instance, we might postulate an elementary tree for the verb {\em
arrived\/} that includes a substitution node for a fronted adverbial
{\em Wh\/} phrase.} One natural analysis of the sentence
\enumsentence{
At what time did Brockway say Harrison arrived?
}
would involve adjunction of a predicative tree for the
phrase ``did Brockway say'' at the root of the tree for ``Harrison
arrived''.  A {\em Wh\/} modifier tree ``at what time'' must be adjoined
in as well.  The example question is ambiguous, of course, as to
whether it questions the time of the saying or of the arriving.  In
the former case, the modifier tree presumably adjoins at the root of
the predicative tree for ``did Brockway say'' that it modifies.  In
the latter case, which is of primary interest here, it must adjoin at
the root of the tree for ``Harrison arrived''.  Thus, both trees would
be adjoined at the same address, and the outermost-predication rule
would predict the derived sentence to be ``Did Brockway say at what
time Harrison arrived.''  To get around this problem, we might specify
hard ordering constraints for English that place all {\em Wh\/} modifier
trees after all predicative trees, which in turn come after all
non-{\em Wh\/} modifier trees.  This would place the {\em Wh\/} modifier
outermost as required.

Although we find this extra flexibility to be an attractive aspect of
this approach, we stay with the more stringent outermost-predication
restriction in the material that follows.

\section{Compilation of TAGs to Linear Indexed Grammars}

In this section, we present a technique for compiling tree-adjoining
grammars into linear indexed grammars such that the linear-indexed
grammar makes explicit the extended derivations of the TAG.  This
compilation plays two roles.  First, it provides for a simple proof of
the generative equivalence of TAGs under the standard and extended
definitions of derivation, as described at the end of this section.
Second, it can be used as the basis for a parsing algorithm that
recovers the extended derivations for strings.  The design of such an
algorithm is the topic of Section~\ref{sec:parse}.

Linear indexed grammars (LIG) constitute a grammatical framework based,
like context-free, context-sensitive, and unrestricted rewriting
systems, on rewriting strings of nonterminal and terminal symbols.
Unlike these systems, linear indexed grammars, like the indexed grammars
from which they are restricted, allow stacks of marker symbols, called
{\em indices}, to be associated with the nonterminal symbols being
rewritten.  The linear version of the formalism allows the full index
information from the parent to be used to specify the index information
for only one of the child constituents.  Thus, a linear indexed
production can be given schematically as:

\[ N_0[..\beta_0] \ra N_1[\beta_1] \cdots N_{s-1}[\beta_{s-1}]\; N_s[..\beta_s]
\; N_{s+1}[\beta_{s+1}] \cdots N_k[\beta_k] \]

\noindent The $N_i$ are nonterminals, the $\beta_i$ strings of
indices.  The {``..''} notation stands for the remainder of the stack
below the given string of indices.  Note that only one element on the
right-hand side, $N_s$, inherits the remainder of the stack from the
parent.  (This schematic  rule is intended to be indicative, not
definitive.  We ignore issues such as the optionality of the inherited
stack, how terminal symbols fit in, and so forth.  Vijay-Shanker and
Weir \shortcite{vw90} present a complete discussion.)

Vijay-Shanker and Weir \shortcite{vw90} present a way of specifying any
TAG as a linear indexed grammar.  The LIG version makes explicit the
standard notion of derivation being presumed.  Also, the LIG version of
a TAG grammar can be used for recognition and parsing.  Because the LIG
formalism is based on augmented rewriting, the parsing algorithms can be
much simpler to understand and easier to modify, and no loss of
generality is incurred.  For these reasons, we use the technique in this
work.

\begin{figure*}
{\qbox{\psfig{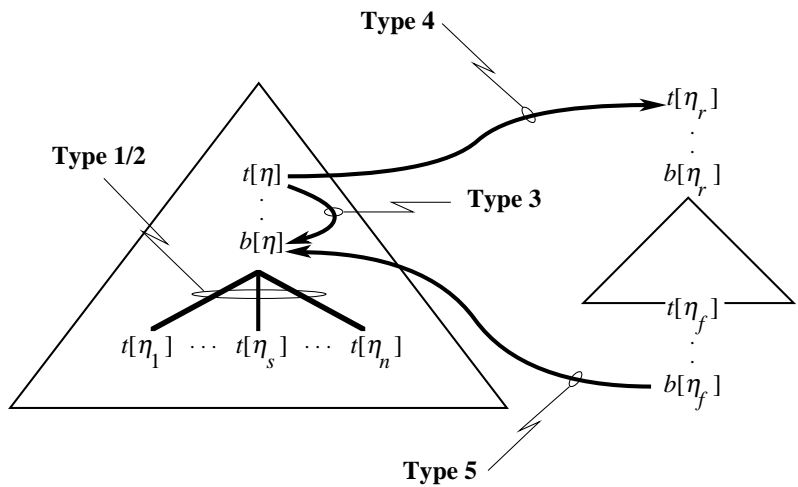}}}
\caption{Schematic structure of adjunction with top and bottom of each
node separated}
\label{fig:adjoin-fig}
\end{figure*}

The compilation process that manifests the standard definition of
derivation can be most easily understood by viewing nodes in a TAG
elementary tree as having both a top and bottom component, identically
marked for nonterminal category, that dominate (but may not immediately
dominate) each other.  (See Figure~\ref{fig:adjoin-fig}.)  The rewrite
rules of the corresponding linear indexed grammar capture the immediate
domination between a bottom node and its child top nodes directly, and
capture the domination between top and bottom parts of the same node by
optionally allowing rewriting from the top of a node to an appropriate
auxiliary tree, and from the foot of the auxiliary tree back to the
bottom of the node.  The index stack keeps track of the nodes that
adjunction has occurred on so that the recognition to the left and the
right of the foot node will occur under identical assumption of
derivation structure.

\begin{figure}
{\qbox{\psfig{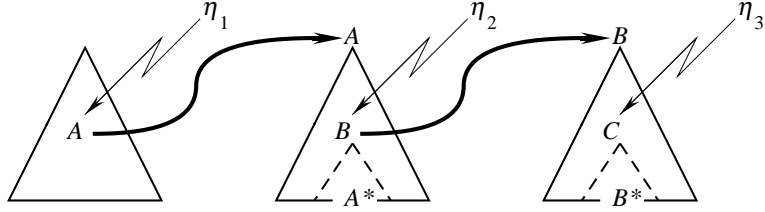}}}
\caption{A stack of indices $[\eta_1 \eta_2 \eta_3]$ captures the
adjunction history that led to the reaching of the node $\eta_3$ in
the parsing process.  Parsing of an elementary tree $\alpha$ proceeded
to node $\eta_1$ in that tree, at which point adjunction of the tree
containing $\eta_2$ was pursued by the parser.  When the node $\eta_2$
was reached, the tree containing $\eta_3$ was implicitly adjoined.
Once this latter tree is completely parsed, the remainder of the tree
containing $\eta_2$ can be parsed from that point, and so on.}
\label{fig:stacks}
\end{figure}

The TAG grammar is encoded as a LIG with two nonterminal symbols $t$
and $b$ corresponding to the top and bottom components, respectively,
of each node.  The stack indices correspond to the individual nodes of
the elementary trees of the TAG grammar.  Thus, there are as many
stack index symbols as there are nodes in the elementary trees of the
grammar, and each such index (i.e., node) corresponds unambiguously to
a single address in a single elementary tree.  (In fact, the symbols
can be thought of as pairs of an elementary tree identifier and an
address within that tree, and our implementation encodes them in just
that way.)  The index at the top of the stack corresponds to the node
being rewritten.  Thus, a LIG nonterminal with stack $t[\eta]$
corresponds to the top component of node $\eta$, and
$b[\eta_1\eta_2\eta_3]$ corresponds to the bottom component of
$\eta_3$.  The indices $\eta_1$ and $\eta_2$ capture the history of
adjunctions that are pending completion of the tree in which $\eta_3$
is a node.  Figure~\ref{fig:stacks} depicts the interpretation of a
stack of indices.

In summary, given a tree-adjoining grammar, the following LIG rules
are generated:
\begin{enumerate}

\item {\em Immediate domination dominating foot:} For each auxiliary
tree node $\eta$ that dominates the foot node, with children $\eta_1,
\ldots, \eta_s, \ldots, \eta_n$, where $\eta_s$ is the child that also
dominates the foot node, include a production

\[ b[..\eta] \ra t[\eta_1] \cdots t[\eta_{s-1}] t[..\eta_s]
t[\eta_{s+1}] \cdots t[\eta_n] \]

\item {\em Immediate domination not including foot:} For each
elementary tree node $\eta$ that does not dominate a foot node, with
children $\eta_1, \ldots, \eta_n$, include a production

\[ b[\eta] \ra t[\eta_1] \cdots t[\eta_n] \]

\item {\em No adjunction:} For each elementary tree node $\eta$ that
is not marked for substitution or obligatory adjunction, include a
production

\[ t[..\eta] \ra b[..\eta] \]

\item {\em Start root of adjunction:} For each elementary tree node
$\eta$ on which the auxiliary tree $\beta$ with root node $\eta_r$ can
be adjoined, include the following production:

\[ t[..\eta] \ra t[..\eta\eta_r] \]

\item {\em Start foot of adjunction:} For each elementary tree node
$\eta$ on which the auxiliary tree $\beta$ with foot node $\eta_f$ can
be adjoined, include the following production:

\[ b[..\eta\eta_f] \ra b[..\eta] \]

\item {\em Start substitution:} For each elementary tree node $\eta$
marked for substitution on which the initial tree $\alpha$ with root
node $\eta_r$ can be substituted, include the production

\[ t[\eta] \ra t[\eta_r] \]

\end{enumerate}

We will refer to productions generated by Rule $i$ above as Type $i$
productions.  For example, Type 3 productions are of the form \(
t[..\eta] \ra b[..\eta] \).  For further information concerning the
compilation see the work of Vijay-Shanker and Weir \shortcite{vw90}.
For present purposes, it is sufficient to note that the method
directly embeds the standard notion of derivation in the rewriting
process.  To perform an adjunction, we move (by Rule 4) from the node
adjoined at to the top of the root of the auxiliary tree.  At the
root, additional adjunctions might be performed.  When returning from
the foot of the auxiliary tree back to the node where adjunction
occurred, rewriting continues at the bottom of the node (see Rule 5),
not the top, so that no more adjunctions can be started at that node.
Thus, the dependent nature of predicative adjunction is enforced
because {\em only a single adjunction can occur at any given node}.

\begin{figure}
{\qbox{\psfig{figure=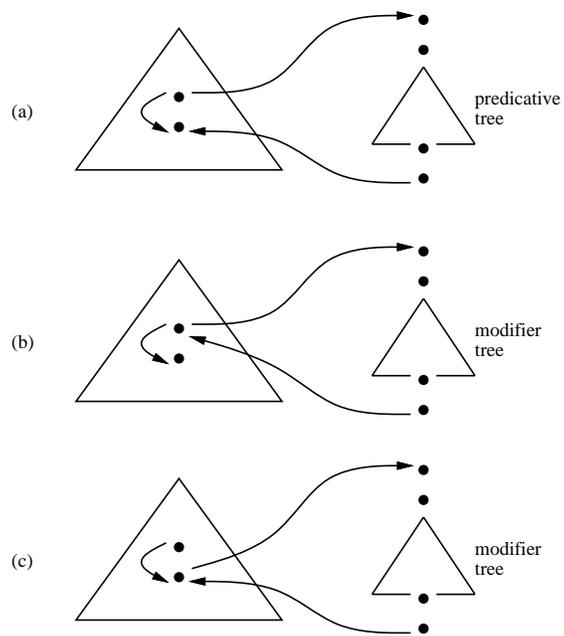,scale=80}}}
\caption{Schematic structure of possible predicative and modifier
adjunctions with top and bottom of each node separated.}
\label{fig:cycles}
\end{figure}

In order to permit extended derivations, we must allow for multiple
modifier tree adjunctions at a single node.  There are two natural ways
this might be accomplished, as depicted in Figure~\ref{fig:cycles}.
\begin{enumerate}

\item {\em Modified start foot of adjunction rule:}  Allow moving from
the bottom of the foot of a modifier auxiliary tree to the top (rather
than the bottom) of the node at which it adjoined (Figure~\ref{fig:cycles}b).

\item {\em Modified start root of adjunction rule:}  Allow moving from
the bottom (rather than the top) of a node to the top of the root of a
modifier auxiliary tree (Figure~\ref{fig:cycles}c).

\end{enumerate}

\noindent As can be seen from the figures, both of these methods allow
recursion at a node, unlike the original method depicted in
Figure~\ref{fig:cycles}a.  Thus multiple modifier trees are allowed to
adjoin at a single node.  Note that since predicative trees fall under
the original rules, at most a single predicative tree can be adjoined
at a node.  The two methods correspond exactly to the innermost- and
outermost-predication methods discussed in
Section~\ref{sec:inner-outer}.  For the reasons described there, the
latter is preferred.  \footnote{The more general definition allowing
predicative trees to occur anywhere within a sequence of modifier
adjunctions would be achieved by adding both types of rules.}

In summary, independent derivation structures can be allowed for
modifier auxiliary trees by starting the adjunction process from the
bottom, rather than the top of a node for those trees.  Thus, we split
Type 4 LIG productions into two subtypes for predicative and modifier
trees, respectively.
\begin{enumerate}

\item[4a.] {\em Start root of predicative adjunction:} For each
elementary tree node $\eta$ on which the predicative auxiliary tree
$\beta$ with root node $\eta_r$ can be adjoined, include the following
production:

\[ t[..\eta] \ra t[..\eta\eta_r] \]

\item[4b.] {\em Start root of modifier adjunction:} For each elementary tree
node $\eta$ on which the modifier auxiliary tree $\beta$ with root node
$\eta_r$ can be adjoined, include the following production:

\[ b[..\eta] \ra t[..\eta\eta_r] \]

\end{enumerate}
Once this augmentation has been made, we no longer need to
allow for adjunctions at the root nodes of modifier auxiliary trees,
as repeated adjunction is now allowed for by the new rule 4b.
Consequently, grammars should forbid adjunction of a modifier tree
$\beta_1$ at the root of a modifier tree $\beta_2$ except where
$\beta_1$ is intended to modify $\beta_2$ directly.

This simple modification to the compilation process from TAG to LIG
fully specifies the modified notion of derivation.  Note that the
extra criterion (5) noted in Section~\ref{sec:criteria} is satisfied
by this definition: Modifier adjunctions are inherently repeatable and
eliminable as the movement through the adjunction ``loop'' ends up at
the same point that it begins.  The recognition algorithms for TAG
based on this compilation, however, must be adjusted to allow for the
new rule types.

This compilation makes possible a simple proof of the weak-generative
equivalence of TAGs under the standard and extended
derivations.\footnote{We are grateful to K. Vijay-Shanker for bringing
this point to our attention.} Call the set of languages generable by a
TAG under the standard definition of derivation $TAL_s$ and under the
extended definition $TAL_e$.  Clearly, $TAL_s \subseteq TAL_e$ since
the standard definition can be mimicked by making all auxiliary trees
predicative.  The compilation above provides the inclusion $TAL_e
\subseteq LIL$, where $LIL$ is the set of linear indexed languages.
The final inclusion $LIL \subseteq TAL_s$ has been shown indirectly by
Vijay-Shanker \shortcite{v87} using embedded push-down automata and
modified head grammars as intermediaries.  From these inclusions, we
can conclude that $TAL_s = TAL_e$.

\section{Recognition and Parsing}
\label{sec:parse}

A recognition algorithm for TAGs can be constructed based on the above
translation into corresponding LIGs as specified by Rules 1 through 6
in the previous section.  The algorithm is not a full recognition
algorithm for LIGs, but rather, is tuned for exactly the types of
rules generated as output of this compilation process.  In this
section, we present the recognition algorithm and modify it to work
with the extended derivation compilation.

We will use the following notations in this and later sections.  The
symbol $P$ will serve as a variable over the two LIG grammar
nonterminals $t$ and $b$.  The substring of the string $w_1
\cdots w_n$ being parsed between indices $i$ and $j$ will be notated
as $w_{i+1} \cdots w_j$, which we take to be the empty string when $i$
is greater than or equal to $j$.  We will use $\Gamma$, $\Delta$, and
$\Theta$ for sequences containing terminals and LIG nonterminals with
their stack specifications.  For instance, $\Gamma$ might be
$t[\eta_1] t[..\eta_2] t[\eta_3]$.

The parsing algorithm can be seen as a tabular parsing method
based on deduction of items, as in Earley deduction
\cite{pereira-warren-p-as-d}.  We will so describe it, by presenting
inference rules over items of the form
\[\eitem{P[\eta]}{\Gamma}{\Delta}{i}{j}{k}{l} \]
Such items play the role of the items of Earley's algorithm.  Unlike
the items of Earley's algorithm, however, an item of this form does
not embed a grammar rule proper, that is \(P[\eta] \ra \Gamma\Delta\)
is not necessarily a rule of the grammar.  Rather, it is what we will
call a {\em reduced\/} rule; for reasons described below, the
nonterminals in $\Gamma$ and $\Delta$ as well as the nonterminal
$P[\eta]$ record only the top element of each stack of indices.  We
will use the notation \(\unreducedra{P[\eta]}{\Gamma \Delta}\) for the
unreduced form of the rule whose reduced form is \(P[\eta] \ra
\Gamma\Delta\).  For instance, the rule specified by the notation
\(\unreducedra{t[\eta_1]}{t[\eta_2]}\) might be the rule
\(t[..\eta_1] \ra t[..\eta_1\eta_2]\).  The reader can easily verify
that the TAG to LIG compilation is such that there is a one-to-one
correspondence between the generated rules and their reduced form.
Consequently, this notation is well-defined.

The dot in the items is analogous to that found in Earley and LR items
as well.  It serves as a marker for how far recognition has proceeded
in identifying the subconstituents for this rule.  The indices $i$,
$j$, $k$, and $l$ specify the portion of the string $w_1
\cdots w_n$ covered by the recognition of the item.  The substring
between $i$ and $l$ (i.e., $w_{i+1} \cdots w_l$) has been recognized,
perhaps with a region between $j$ and $k$ where the foot of the tree
below the node $\eta$ has been recognized.  (If the foot node is not
dominated by $\Gamma$, we take the values of $j$ and $k$ to be the
dummy value `$-$'.)

\subsection{The Inference Rules}

In this section, we specify several inference rules for parsing a LIG
generated from a TAG, which we recall in this section.  One
explanatory comment is in order, however, before the rules are
presented.  The rules of a LIG associate with each constituent a
nonterminal and a stack of indices.  It seems natural for a parsing
algorithm to maintain this association by building items that specify
for each constituent the full information of nonterminal and index
stack.  However, this would necessitate storing an unbounded amount of
information for each potential constituent, resulting in a parsing
algorithm that is potentially quite inefficient when nondeterminism
arises during the parsing process, and perhaps non-effective if the
grammar is infinitely ambiguous.  Instead, the parse items manipulated
by the inference rules that we present do not keep all of this
information for each constituent.  Rather, the items keep only the
single top stack element for each constituent (in addition to the
nonterminal symbol).  This drastically decreases the number of
possible items, and accounts for the polynomial character of the
resultant algorithm.\footnote{Vijay-Shanker and Weir \shortcite{vw90} first
proposed the recording of only the top stack element in order to
achieve efficient parsing.  The algorithm they presented is a
bottom-up general LIG parsing algorithm.  Schabes
\shortcite{schabes91c} sketches a proof of an $O(n^6)$ bound for an
Earley-style algorithm for TAG parsing that is more closely related to
the algorithm proposed here.} Side conditions make up for some of the
loss of information, thereby maintaining correctness.  For instance,
the Type 4 Completor rule specifies a relation between $\eta$ and
$\eta_f$ that takes the place of popping an element off of the stack
associated with $\eta$.  However, the side conditions are strictly
weaker than maintaining full stack information.  Consequently, the
algorithm, though correct, does not maintain the valid prefix
property.  See the work of Schabes
\shortcite{schabes91c} for further discussion and alternatives.

Scanning and prediction work much as in Earley's original algorithm.
\begin{itemize}

\item {\em Scanner:}

\[ \oneover{   \eitem{b[\eta]}{\Gamma}{a \Delta}{i}{j}{k}{l}    }
           {   \eitem{b[\eta]}{\Gamma a}{\Delta}{i}{j}{k}{l+1}    }
	\qquad a=w_{l+1}\]

Note that the only rules that need be considered are those where the
parent is a bottom node, as terminal symbols occur on the right-hand
side only of Type 1 or 2 productions.  Otherwise, the rule is exactly
as that for Earley's algorithm except that the extra foot indices ($j$
and $k$) are carried along.

\item {\em Predictor:}

\[ \oneover{   \eitem{P[\eta]}{\Gamma}{P'[\eta'] \Delta}{i}{j}{k}{l}
}
           {   \eitem{P'[\eta']}{ }{\Theta}{l}{-}{-}{l}
}
	\qquad \unreducedra{P'[\eta']}{\Theta} \]

This rule serves to form predictions for any type production in the
grammar, as the variables $P$ and $P'$ range over the values $t$ and
$b$.  In the predicted item, the foot is not dominated by the (empty)
recognized input, so that the dummy value `$-$' is used for the foot
indices.  Note that the predicted item records the reduced form of an
unreduced rule \(\unreducedra{P'[\eta']}{\Theta}\) of the grammar.

\end{itemize}

Completion of items (moving of the dot from left to right over a
nonterminal) breaks up into several cases, depending on which
production type is being completed.  This is because the addition of
the extra indices and the separate interpretations for top and bottom
productions require differing index manipulations to be performed.  We
will list the various steps, organized by what type of production they
participate in the completion of.

Productions that specify immediate domination (from Rules 1 and 2) are
completed whenever the top of the child node is fully recognized.
\begin{itemize}

\item {\em Type 1 and 2 Completor:}

\[ \oneover{   	\eitem{b[\eta_1]}{\Gamma}{t[\eta] \Delta}{m}{j'}{k'}{i}
			\quad
		\eitem{t[\eta]}{\Theta}{ }{i}{j}{k}{l}
  }
           {   \eitem{b[\eta_1]}{\Gamma t[\eta]}{\Delta}{m}
                    {j \cup j'}{k \cup k'}{l}
 }
\]

Here, $t[\eta]$ has been fully recognized as the substring between $i$
and $l$.  The item expecting $t[\eta]$ can be completed.  One of the
two antecedent items might also dominate the foot node of the tree to
which $\eta$ and $\eta_1$ belong, and would therefore have indices for
the foot substring.  The operations $j \cup j'$ and $k \cup k'$ are
used to specify whichever of $j$ or $j'$ (and respectively for $k$ or
$k'$) contain foot substring indices.  The formal definition of $\cup$
is as follows:

\[j \cup j' = \left\{ \begin{array}{ll}
      j  & \mbox{ if $j' = -$} \\
      j' & \mbox{ if $j  = -$} \\
      j  & \mbox{ if $j' = j$} \\
      \mbox{undefined}  & \mbox{ otherwise}
    \end{array} \right. \]

\end{itemize}

The remaining rules (3 through 6) are each completed by a particular
completion instance.

\begin{itemize}

\item {\em Type 3 Completor:}

\[ \oneover{   \eitem{t[\eta]}{ }{b[\eta]}{i}{-}{-}{i}     \quad
	       \eitem{b[\eta]}{\Theta}{ }{i}{j}{k}{l}          }
           {   \eitem{t[\eta]}{b[\eta]}{ }{i}{j}{k}{l}         }
\]

This rule is used to complete a prediction that no (predicative)
adjunction occurs at node $\eta$.  Once the part of the string
dominated by $b[\eta]$ has been found, as evidenced by the second
antecedent item, the prediction of no adjunction can be completed.

\item {\em Type 4 Completor:}

\[ \oneover{  \begin{array}{l}
		 \eitem{t[\eta]}{ }{t[\eta_r]}{i}{-}{-}{i} \\
          \itemoffset \eitem{t[\eta_r]}{\Theta}{ }{i}{j}{k}{l}\\
   \itemoffset \itemoffset \eitem{b[\eta]}{\Delta}{ }{j}{p}{q}{k}
	      \end{array}    }
           {   \eitem{t[\eta]}{t[\eta_r]}{ }{i}{p}{q}{l}     }
	\qquad t[..\eta] \ra t[..\eta\eta_r] \]

Here, an adjunction has been predicted at $\eta$, and the adjoined
derived tree (between $t[\eta]$ and $b[\eta]$) and the derived
material that $\eta$ itself dominates (below $b[\eta]$) have both been
completed.  Thus $t[\eta]$ is completely recognized.  Note that the
side condition (the unreduced form of the reduced rule in the first
antecedent item) is placed merely to guarantee that $\eta_r$ is the
root node of an adjoinable auxiliary tree.

\item {\em Type 5 Completor:}

\[ \oneover{   \eitem{b[\eta_f]}{ }{b[\eta]}{i}{-}{-}{i}     \quad
               \eitem{b[\eta]}{\Theta}{ }{i}{j}{k}{l}            }
           {   \eitem{b[\eta_f]}{b[\eta]}{ }{i}{i}{l}{l}         }
	\qquad b[..\eta\eta_f] \ra b[..\eta]\]

When adjunction has been performed, and recognition up to the foot
node $\eta_f$ has been performed, it is necessary to recognize all the
material under the foot node.  When that is done, the foot node
prediction can be completed.  Note that it must be possible to have
adjoined the auxiliary tree at node $\eta$ as specified in the
production in the side condition.

\item {\em Type 6 Completor:}

\[ \oneover{   \eitem{t[\eta]}{ }{t[\eta_r]}{i}{-}{-}{i}   \quad
               \eitem{t[\eta_r]}{\Theta}{ }{i}{-}{-}{l}        }
           {   \eitem{t[\eta]}{t[\eta_r]}{ }{i}{-}{-}{l}       }
	\qquad t[\eta] \ra t[\eta_r]\]

Completion of the material below the root node $\eta_r$ of an initial
tree allows for the completion of the node at which substitution
occurred.

\end{itemize}

\begin{sloppypar}
The recognition process for a string $w_1 \cdots w_n$ starts with some
items that serve as axioms for these inference rules.  For each rule
$\unreducedra{t[\eta_s]}{\Gamma}$ where $\eta_s$ is the root node of an
initial tree which node is labeled with the start nonterminal, the
item $\eitem{t[\eta_s]}{ }{\Gamma}{0}{-}{-}{0}$ is an axiom.  If from
these axioms an item of the form $\eitem{t[\eta_s]}{\Gamma}{
}{0}{-}{-}{n}$ can be proved according to the rules of inference
above, the string is accepted; otherwise it is rejected.
\end{sloppypar}

Alternatively, the axioms can be stated as if there were extra rules
$S\ra t[\eta_s]$ for each $\eta_s$ a start-nonterminal-labeled root
node of an initial tree.  In this case, the axioms are items of the
form $\eitem{S}{ }{t[\eta_s]}{0}{-}{-}{0}$ and the string is accepted
upon proving $\eitem{S}{t[\eta_s]}{ }{0}{-}{-}{n}$.  In this case, an
extra prediction and completion rule is needed just for these rules,
since the normal rules do not allow $S$ on the left-hand side.  This
point is taken up further in Section~\ref{sec:deriv-axioms}.

Generation of items can be cached in the standard way for
inference-based parsing algorithms \cite{shieber-thesis}; this leads
to a tabular or chart-based parsing algorithm. \label{sec:axioms}

\subsection{The Algorithm Invariant}

The algorithm maintains an invariant that holds of all items added to
the chart.  We will describe the invariant using some additional
notational conventions.  Recall that \(\unreducedra{P[\eta]}{\Gamma}\)
is the LIG production in the grammar whose reduced form is \({P[\eta]}
\ra {\Gamma}\).  The notation $\Gamma[\gamma]$ where
$\gamma$ is a sequence of stack symbols (i.e., nodes), specifies the
sequence $\Gamma$ with $\gamma$ replacing the occurrence of $..$ in
the stack specifications.  For example, if $\Gamma$ is the sequence
$t[\eta_1] t[..\eta_2] t[\eta_3]$, then $\Gamma[\gamma] = t[\eta_1]
t[\gamma\eta_2] t[\eta_3]$.  A single LIG derivation step will be
notated with $\Ra$ and its reflexive transitive closure with $\Ra^*$.

The invariant specifies that
$\eitem{P[\eta]}{\Gamma}{\Delta}{i}{j}{k}{l}$ is in the chart only
if\footnote{The invariant is not stated as a biconditional because
this would require strengthening of the antecedent condition.  The
natural strengthening, following the standard for Earley's algorithm,
would be to add a requirement that the item be consistent with left
context, as
\begin{enumerate}
\item[(d)] \qquad $\eta_s \Ra^* w_1 \cdots w_i P[\gamma\eta]$
\end{enumerate}
but this is too strong.  This condition implies that the algorithm
possesses the valid prefix property, which it does not.  The exact
statement of the invariant condition that would allow for exact
specifications of the item semantics is the topic of ongoing research.
However, the current specification is sufficient for proving soundness
of the algorithm.}
\begin{enumerate}

\item If node $\eta$ dominates the foot node $\eta_f$ of the tree to which
it belongs, then there exists a string of stack symbols (i.e., nodes)
$\gamma$ such that

\begin{enumerate}

\item $\unreducedra{P[\eta]}{\Gamma \Delta}$ is a LIG rule in the
grammar, where $\unreduced{\Gamma}$ is the unreduced form of $\Gamma$.

\item $\unreduced{\Gamma}[\gamma \eta] \Ra^* w_{i+1} \cdots w_j
b[\gamma \eta_f] w_{k+1} \cdots w_l$\eqpunc{.}

\item $b[\gamma \eta_f] \Ra^* w_{j+1} \cdots w_k$\eqpunc{.}

\end{enumerate}

\item If node $\eta$ does not dominate the foot node $\eta_f$ of the
tree to which it belongs or there is no foot node in the tree, then

\begin{enumerate}

\item $\unreducedra{P[\eta]}{\Gamma \Delta}$ is a LIG rule in the
grammar, where $\unreduced{\Gamma}$ is the unreduced form of $\Gamma$.

\item $\unreduced{\Gamma} \Ra^* w_{i+1} \cdots w_l$\eqpunc{.}

\item $j$ and $k$ are not bound.

\end{enumerate}

\end{enumerate}

According to this invariant, for a node $\eta_s$ which is the root of
an initial tree, the item $\eitem{t[\eta_s]}{\Gamma}{ }{0}{-}{-}{n}$
is in the chart only if $t[\eta_s] \Ra \unreduced{\Gamma} \Ra^* w_1
\cdots w_n$.  Thus, soundness of the algorithm as a recognizer
follows.

\subsection{Modifications for Extended Derivations} \label{sec:inf-extend}

Extending the algorithm to allow for the new types of production
(specifically, as derived by Rule 4b) requires adding a completion
rule for Type 4b productions.  For the new type of production, a
completion rule of the following form is required:
\begin{itemize}

\item {\em Type 4b Completor:}

\[ \oneover{  \begin{array}{l}
		 \eitem{b[\eta]}{ }{t[\eta_r]}{i}{-}{-}{i} \\
          \itemoffset \eitem{t[\eta_r]}{\Theta}{ }{i}{j}{k}{l}\\
   \itemoffset \itemoffset \eitem{b[\eta]}{\Delta}{ }{j}{p}{q}{k}
	      \end{array}    }
           {   \eitem{b[\eta]}{t[\eta_r]}{ }{i}{p}{q}{l}     }
	\qquad b[..\eta] \ra t[..\eta\eta_r] \]

\end{itemize}

In addition to being able to complete Type 4b items, we must also be
able to complete other items using completed Type 4b items.  This is
an issue in particular for completor rules that might move their dot
over a $b[\eta]$ constituent, in particular, the Type 3 and 5
Completors.  However, these rules have been stated so that the
antecedent item with right-hand side $b[\eta]$ already matches Type 4b
items.  Furthermore, the general statement, including index
manipulation is still appropriate in the context of Type 4b
productions.  Thus, no further changes to the recognition inference
rules are needed for this purpose.

However, a bit of care must be taken in the interpretation of the Type
{1/2} Completor.  Type 4b items that require completion bear a
superficial resemblance to Type 1 and 2 items, in that both have a
constituent of the form $t[\dumarg]$ after the dot.  In Type 4b items,
the constituent is $t[\eta_r]$, in Type 4a items $t[\eta]$.  But it is
crucial that the Type {1/2} Completor not be used to complete Type 4b
items.  A simple distinguishing characteristic is that in Type 1 and 2
items to be completed, the node $\eta$ after the dot is never a root
node (as it is immediately dominated by $\eta_1$), whereas in Type 4b
items, the node $\eta_r$ after the dot is always a root node (of a
modifier tree).  Simple side conditions can distinguish the cases.

Figure~\ref{fig:all-rules} contains the final versions of the inference
rules for recognition of LIGs corresponding to extended TAG
derivations.

\begin{figure*}
{\small\begin{itemize}

\item {\em Scanner:}

\[ \oneover{   \eitem{b[\eta]}{\Gamma}{a \Delta}{i}{j}{k}{l}    }
           {   \eitem{b[\eta]}{\Gamma a}{\Delta}{i}{j}{k}{l+1}    }
	\qquad a=w_{l+1}\]

\item {\em Predictor:}

\[ \oneover{   \eitem{P[\eta]}{\Gamma}{P'[\eta'] \Delta}{i}{j}{k}{l}
}
           {   \eitem{P'[\eta']}{ }{\Theta}{l}{-}{-}{l}
}
	\qquad \unreducedra{P'[\eta']}{\Theta} \]

\item {\em Type 1 and 2 Completor:}

\[ \oneover{   	\eitem{b[\eta_1]}{\Gamma}{t[\eta] \Delta}{m}{j'}{k'}{i}
			\quad
		\eitem{t[\eta]}{\Theta}{ }{i}{j}{k}{l}
  }
           {   \eitem{b[\eta_1]}{\Gamma t[\eta]}{\Delta}{m}
                    {j \cup j'}{k \cup k'}{l}
 } \qquad \mbox{$\eta$ not a root node}
\]

\item {\em Type 3 Completor:}

\[ \oneover{   \eitem{t[\eta]}{ }{b[\eta]}{i}{-}{-}{i}     \quad
	       \eitem{b[\eta]}{\Theta}{ }{i}{j}{k}{l}          }
           {   \eitem{t[\eta]}{b[\eta]}{ }{i}{j}{k}{l}         }
\]

\item {\em Type 4a Completor:}

\[ \oneover{  \begin{array}{l}
		 \eitem{t[\eta]}{ }{t[\eta_r]}{i}{-}{-}{i} \\
          \itemoffset \eitem{t[\eta_r]}{\Theta}{ }{i}{j}{k}{l}\\
   \itemoffset \itemoffset \eitem{b[\eta]}{\Delta}{ }{j}{p}{q}{k}
	      \end{array}    }
           {   \eitem{t[\eta]}{t[\eta_r]}{ }{i}{p}{q}{l}     }
	\qquad t[..\eta] \ra t[..\eta\eta_r] \]

\item {\em Type 4b Completor:}

\[ \oneover{  \begin{array}{l}
		 \eitem{b[\eta]}{ }{t[\eta_r]}{i}{-}{-}{i} \\
          \itemoffset \eitem{t[\eta_r]}{\Theta}{ }{i}{j}{k}{l}\\
   \itemoffset \itemoffset \eitem{b[\eta]}{\Delta}{ }{j}{p}{q}{k}
	      \end{array}    }
           {   \eitem{b[\eta]}{t[\eta_r]}{ }{i}{p}{q}{l}     }
	\qquad b[..\eta] \ra t[..\eta\eta_r] \]

\item {\em Type 5 Completor:}

\[ \oneover{   \eitem{b[\eta_f]}{ }{b[\eta]}{i}{-}{-}{i}     \quad
               \eitem{b[\eta]}{\Theta}{ }{i}{j}{k}{l}            }
           {   \eitem{b[\eta_f]}{b[\eta]}{ }{i}{i}{l}{l}         }
	\qquad b[..\eta\eta_f] \ra b[..\eta]\]

\item {\em Type 6 Completor:}

\[ \oneover{   \eitem{t[\eta]}{ }{t[\eta_r]}{i}{-}{-}{i}   \quad
               \eitem{t[\eta_r]}{\Theta}{ }{i}{-}{-}{l}        }
           {   \eitem{t[\eta]}{t[\eta_r]}{ }{i}{-}{-}{l}       }
	\qquad t[\eta] \ra t[\eta_r]\]

\end{itemize}}
\caption{Inference Rules for Extended Derivation TAG Recognition}
\label{fig:all-rules}
\end{figure*}

\subsection{Maintaining Derivation Structures}

One of the intended applications for extended derivation TAG parsing is
the parsing of synchronous TAGs.  Especially important in this
application is the ability to generate the derivation trees while
parsing proceeds.

A synchronous TAG is composed of two base TAGs (which we will call the
{\em source\/} and {\em target\/} TAG) whose elementary trees have been
paired one-to-one.  A synchronous TAG whose source TAG is a grammar
for a fragment of English, and whose target TAG is a grammar for a
logical form language may be used to generate logical forms for each
sentence of English that the source grammar admits
\cite{shieber-schabes}.  Similarly, with source and target swapped,
the synchronized grammar may be used to generate English sentences
corresponding to logical forms \cite{ss91}.  If the source and target
grammars specify fragments of natural languages, an automatic
translation system is specified \cite{asj90}.

Abstractly viewed, the processing of a synchronous grammar proceeds
by parsing an input string according to the source grammar, thereby
generating a derivation tree for the string; mapping the derivation
tree into a derivation tree for the target grammar; and generating a
derived tree (hence, derived string) according to the target grammar.

One frequent worry about synchronous TAGs as used in their semantic
interpretation mode is whether it is possible to perform incremental
interpretation.  The abstract view of processing just presented seems
to require that a full derivation tree be developed before
interpretation into the logical form language can proceed.
Incremental interpretation, on the other hand, would allow partial
interpretation results to guide the parsing process on-line, thereby
decreasing the nondeterminism in the parsing process.  Whether
incremental interpretation is possible depends precisely on the extent
to which the three abstract phases of synchronous TAG processing can
in fact be interleaved.  In previous work, we left this issue open.
In this section, we allay these worries by showing how the extended
TAG parser just presented can build derivation trees incrementally as
parsing proceeds.  Once this has been demonstrated, it should be
obvious that these derivation trees could be transferred to target
derivation trees during the parsing process, and immediately generated
from.  Thus, incremental interpretation is demonstrated to be possible
in the synchronous TAG framework.  In fact, the technique presented in
this section has allowed for the first implementation of synchronous
TAG processing, due to Onnig Dombalagian.  This implementation was
directly based on the inference-based TAG parser mentioned in
Section~\ref{sec:complexity} and presented in full elsewhere
\cite{full-alt-deriv}.

We associate with each item a set of {\em operations\/} that have been
implicitly carried out by the parser in recognizing the substring
covered by the item.  An operation can be characterized by a
derivation tree and a tree address at which the derivation tree is to
be placed; it corresponds roughly to a branch of a derivation tree.
Prediction items have the empty set of operations.  Type 4 and 6
completion steps build new elements of the sets as they correspond to
actually carrying out adjunction and substitution operations,
respectively.  Other completion steps merely pool the operations from
their constituent parts.

In describing the building of derivation trees, we will use normal set
notation for the sets of derivation trees.  We will assume that for
each node $\eta$, there are functions $tree(\eta)$ and $addr(\eta)$
that specify, respectively, the initial tree that $\eta$\/ occurs in
and its address in that tree.  Finally, we will use a constructor
function for derivation trees $deriv(\gamma, S)$, where $\gamma$
specifies an elementary tree and $S$ specifies a set of operations on
it.  An operation is built with $op(t, D)$ where $t$ is a tree address
and $D$ is a derivation tree to be operated at that address.

Figure~\ref{fig:deriv-rules} lists the previously presented
recognition rules augmented to build derivation structures as the
final component of each item.  The axioms for this inference system
are items of the form $\ditem{S}{ }{t[\eta_s]}{0}{-}{-}{0}{\set{}}$,
where we assume as in Section~\ref{sec:axioms} that there are extra
rules $S \ra t[\eta_s]$ for each $\eta_s$ a start-nonterminal-labeled
root node of an initial tree.  We require an extra rule for prediction
and completion to handle this new type of rule.  The predictor rule is
the obvious analog:
\begin{itemize}
\item {\em Start Rule Predictor:}

\[ \oneover{   \ditem{S}{\Gamma}{P'[\eta'] \Delta}{i}{j}{k}{l}{S}}
           {   \ditem{P'[\eta']}{ }{\Theta}{l}{-}{-}{l}{\set{}}
   }
	\qquad \unreducedra{P'[\eta']}{\Theta} \]
\end{itemize}
In fact, the existing predictor rule could have been easily
generalized to handle this case.

The completor for these start rules is the obvious analog to a Type 6
completor, except in the handling of the derivation.  It delivers,
instead of a set of derivation operations, a single derivation tree.
\begin{itemize}
\item {\em Start Rule Completor:}

\[ \oneover{   \ditem{S}{ }{t[\eta_s]}{i}{-}{-}{i}{\set{}}  \quad
               \ditem{t[\eta_s]}{\Theta}{ }{i}{-}{-}{l}{S}        }
           {   \ditem{S}{t[\eta_s]}{ }{i}{-}{-}{l}
               { deriv(tree(\eta_s), S)}} \]
\end{itemize}

The string is accepted upon proving $\ditem{S}{t[\eta_s]}{
}{0}{-}{-}{n}{D}$, where $D$ is the derivation developed during the
parse.
\label{sec:deriv-axioms}

\begin{figure*}
{\small\begin{itemize}

\item {\em Scanner:}

\[ \oneover{   \ditem{b[\eta]}{\Gamma}{a \Delta}{i}{j}{k}{l}{S}    }
           {   \ditem{b[\eta]}{\Gamma a}{\Delta}{i}{j}{k}{l+1}{S}    }
	\qquad a=w_{l+1}\]

\item {\em Predictor:}

\[ \oneover{   \ditem{P[\eta]}{\Gamma}{P'[\eta'] \Delta}{i}{j}{k}{l}{S}}
           {   \ditem{P'[\eta']}{ }{\Theta}{l}{-}{-}{l}{\set{}}
   }
	\qquad \unreducedra{P'[\eta']}{\Theta} \]

\item {\em Type 1 and 2 Completor:}

\[ \oneover{   \ditem{b[\eta_1]}{\Gamma}{t[\eta] \Delta}
			{m}{j'}{k'}{i}{S_1} \quad
               \ditem{t[\eta]}{\Theta}{ }{i}{j}{k}{l}{S_2}          }
           {   \ditem{b[\eta_1]}{\Gamma t[\eta]}{\Delta}
			{m}{j \cup j'}{k \cup k'}{l}{S_1 \cup S_2}  }
 	\qquad \mbox{$\eta$ not a root node}
\]

\item {\em Type 3 Completor:}

\[ \oneover{   \ditem{t[\eta]}{ }{b[\eta]}{i}{-}{-}{i}{\set{}}    \quad
               \ditem{b[\eta]}{\Theta}{ }{i}{j}{k}{l}{S}    }
           {   \ditem{t[\eta]}{b[\eta]}{ }{i}{j}{k}{l}{S}         }
\]

\item {\em Type 4a Completor:}

\[ \oneover{   \begin{array}{l}
		\ditem{t[\eta]}{ }{t[\eta_r]}{i}{-}{-}{i}{\set{}}	\\
    \itemoffset \ditem{t[\eta_r]}{\Theta}{ }{i}{j}{k}{l}{S_1}   \\
    \itemoffset \itemoffset \ditem{b[\eta]}{\Delta}{ }{j}{p}{q}{k}{S_2}
	       \end{array} }
           {   \ditem{t[\eta]}{t[\eta_r]}{}{i}{p}{q}{l}
               {\set{op(addr(\eta), deriv(tree(\eta_r), S_1))}
			\cup S_2}     }
	\qquad t[..\eta] \ra t[..\eta\eta_r] \]

\item {\em Type 4b Completor:}

\[ \oneover{   \begin{array}{l}
		\ditem{b[\eta]}{ }{t[\eta_r]}{i}{-}{-}{i}{\set{}}	\\
    \itemoffset \ditem{t[\eta_r]}{\Theta}{ }{i}{j}{k}{l}{S_1}   \\
    \itemoffset \itemoffset \ditem{b[\eta]}{\Delta}{ }{j}{p}{q}{k}{S_2}
	       \end{array} }
           {   \ditem{t[\eta]}{t[\eta_r]}{}{i}{p}{q}{l}
               {\set{op(addr(\eta), deriv(tree(\eta_r), S_1))}
			\cup S_2}     }
	\qquad b[..\eta] \ra t[..\eta\eta_r] \]

\item {\em Type 5 Completor:}

\[ \oneover{   \ditem{b[\eta_f]}{ }{b[\eta]}{i}{-}{-}{i}{\set{}}  \quad
               \ditem{b[\eta]}{\Theta}{ }{i}{j}{k}{l}{S}          }
           {   \ditem{b[\eta_f]}{b[\eta]}{ }{i}{i}{l}{l}{S}       }
	\qquad b[..\eta\eta_f] \ra b[..\eta] \]

\item {\em Type 6 Completor:}

\[ \oneover{   \ditem{t[\eta]}{ }{t[\eta_r]}{i}{-}{-}{i}{\set{}}  \quad
               \ditem{t[\eta_r]}{\Theta}{ }{i}{-}{-}{l}{S}        }
           {   \ditem{t[\eta]}{t[\eta_r]}{ }{i}{-}{-}{l}
               {\set{op(addr(\eta), deriv(tree(\eta_r), S))}}     }
	\qquad t[\eta] \ra t[\eta_r] \]
\end{itemize}}
\caption{Inference Rules for Extended Derivation TAG Parsing}
\label{fig:deriv-rules}
\end{figure*}

\subsection{Complexity Considerations}
\label{sec:complexity}

The inference system of Section~\ref{sec:inf-extend} essentially
specifies a parsing algorithm with complexity of $O(n^6)$ in the
length of the string.  Adding explicit derivation structures to the
items, as in the inference system of the previous section eliminates
the polynomial character of the algorithm, in that there may be an
unbounded number of derivations corresponding to any given item of the
original sort.  Even for finitely ambiguous grammars, the number of
derivations may be exponential.  Nonetheless, this fact does not
vitiate the usefulness of the second algorithm, which maintains
derivations explicitly.  The point of this augmentation is to allow
for incremental interpretation --- for interleaved processing of a
post-syntactic sort --- so as to guide the parsing process in making
choices on-line.  By using the extra derivation information, the
parser should be able to eliminate certain nondeterministic paths of
computation; otherwise, there is no reason to do the interpretation
incrementally.  But this determinization of choice presumably
decreases the complexity.  Thus, the extra information is designed for
use in cases where the full search space is not intended to be
explored.

Of course, a polynomial shared-forest representation of the
exponential number of derivations could have been maintained (by
maintaining back pointers among the items in the standard fashion).
For performing incremental interpretation for the purpose of
determinization of parsing, however, the non-shared representation is
sufficient, and preferable on grounds of ease of implementation and
expository convenience.

As a proof of concept, the parsing algorithm just described was
implemented in Prolog on top of a simple, general-purpose,
agenda-based inference engine.  Encodings of explicit inference rules
are essentially interpreted by the inference engine.  The Prolog
database is used as the chart; items not already subsumed by a
previously generated item are asserted to the database as the parser
runs.  An agenda is maintained of potential new items.  Items are
added to the agenda as inference rules are triggered by items added to
the chart.  Because the inference rules are stated explicitly, the
relation between the abstract inference rules described in this paper
and the implementation is extremely transparent.  As a
meta-interpreter, the prototype is not particularly efficient.  (In
particular, the implementation does not achieve the theoretical
$O(n^6)$ bound on complexity, because of a lack of appropriate
indexing.)  Code for the prototype implementation is available for
distribution electronically from the authors.

\section{Conclusion}

The precise formulation of derivation for tree-adjoining grammars has
important ramifications for a wide variety of uses of the formalism,
from syntactic analysis to semantic interpretation and statistical
language modeling.  We have argued that the definition of
tree-adjoining derivation must be reformulated in order to take
greatest advantage of the decoupling of derivation tree and derived
tree by manifesting the proper linguistic dependencies in derivations.
The particular proposal is both precisely characterizable through a
definition of TAG derivations as equivalence classes of ordered
derivation trees, and computationally operational, by virtue of a
compilation to linear indexed grammars together with an efficient
algorithm for recognition and parsing according to the compiled
grammar.

\section*{Acknowledgments}
Order of authors is not intended as an indication of precedence of
authorship.  Much of the work reported in this paper was performed
while the first author was at the Department of Computer and
Information Science, University of Pennsylvania, Philadelphia, PA.
The first author was supported in part by DARPA Grant N0014-90-31863,
ARO Grant DAAL03-89-C-0031 and NSF Grant IRI-90-16592.  The second
author was supported in part by Presidential Young Investigator award
IRI-91-57996 from the National Science Foundation and a matching grant
from Xerox Corporation.  The authors wish to thank Aravind Joshi for
his support of the research, and Aravind Joshi, Judith Klavans,
Anthony Kroch, Shalom Lappin, Kathy McCoy, Fernando Pereira, James
Pustejovsky, and K.~Vijay-Shanker for their helpful discussions of the
issues involved.  We are indebted to David Yarowsky for aid in the
design of the experiment mentioned in
footnote~\protect\ref{fn:experiment} and for its execution.

\bibliographystyle{fullname}

\newpage

\appendix

\section{Proof of Redundancy of Adjacent Sibling Swapping}

\subsection{Preliminaries}

\subsubsection{Tree Addresses}

We define {\em tree addresses\/} (variables over which are
conventionally notated $p, q, \ldots, t, u, v$ and their subscripted
and primed variants) as the finite, possibly empty, sequences of
positive integers (conventionally $i, j, k$), with $\dumarg \conc
\dumarg$ as the sequence concatenation operator.  We uniformly abuse
notation by conflating the distinction between singleton sequences and
their one element.

We use the notation $p \prefix q$ to notate that tree address $p$ is a
proper prefix of $q$, and $p \prefixeq q$ for improper prefix.  When
$p \prefixeq q$, we write $q - p$ for the (possibly empty) sequence
obtained from $q$ by removing $p$ from the front, e.g.\ $1 \conc 2
\conc 3 \conc 4 - 1 \conc 2 = 3 \conc 4$.

\subsubsection{Trees}

We will take trees (conventionally $A, B, E, T$; also $\alpha, \beta,
\gamma$ in the prior text) to be finite partial functions from tree
addresses to symbols, such that the functions are
\begin{description}

\item[Prefix closed:]  For any tree $T$, if $T(p \conc i)$ is defined
then $T(p)$ is defined.

\item[Left closed:]  For any tree $T$, if $T(p \conc i)$ is defined
and $i > 1$ then $T(p \conc (i-1))$ is defined.

\end{description}

We will refer to the domain of a tree $T$, the tree addresses for
which $T$ is defined, as the {\em nodes\/} of $T$.  A node $p$ of $T$ is
a {\em frontier node\/} if $T(p \conc i)$ is undefined for all $i$.  A
node of $T$ is an {\em interior node\/} if it is not a frontier node.
We say that a node $p$ of $T$ is labeled with a symbol $s$ if $T(p) =
s$.

\subsection{Tree-Adjoining Grammars and Derivations}

\subsubsection{Tree-Adjoining Grammars}

In the following definitions, we restrict attention to tree-adjoining
grammars in which adjunction is the only operation; substitution is
not allowed.  The definitions are, however, easily augmented to
include substitution.  We define a {\em tree-adjoining grammar\/} to be
given by a quintuple $\seq{\Sigma, N, \Itree, \Atree, S}$ where
\begin{itemize}

\item $\Sigma$ is a finite set of {\em terminal symbols}.

\item $N$ is a finite set of {\em nonterminal symbols\/} disjoint from
$\Sigma$.

\item ($V = \Sigma \cup N$ is the {\em vocabulary\/} of the grammar.)

\item $S$ is a distinguished nonterminal symbol, the {\em start symbol}.

\item $\Itree$ is a finite set of trees, the {\em initial trees}, where

\begin{itemize}

\item interior nodes are labeled by nonterminal symbols, and

\item frontier nodes are labeled by terminal symbols or the special
symbol $\epsilon$.  (We require that $\epsilon \not \in V$, as $\epsilon$
intuitively specifies the empty string.)

\end{itemize}

\item $\Atree$ is a finite set of trees, the {\em auxiliary trees}, where

\begin{itemize}

\item interior nodes are labeled by nonterminal symbols, and

\item frontier nodes are labeled by terminal symbols or $\epsilon$, except
for one node, called the {\em foot node}, which is labeled with a
nonterminal symbol.

\end{itemize}

\item ($\Etree = \Itree \cup \Atree$ is the set of {\em elementary
trees\/} of the grammar.)

\end{itemize}

By convention, the address of the foot node of a tree $A$ is notated
$f_A$.

\subsubsection{Adjunction}

The adjunction of an auxiliary tree $A$ at address $t$ in tree $E$
notated $E[A/t]$ is defined to be the smallest (least defined) tree
$T$ such that
\[ T(r) = \left\{ \begin{array}{lll}
			E(r) & \mbox{ if $t \not \prefix r$} & (1) \\
			A(u) & \mbox{ if $r = t \conc u$ and $f_A
					\not \prefix u$} & (2)\\
			E(t \conc u) & \mbox{ if $r = t \conc f_A
					\conc u$} & (3)
		  \end{array} \right.
\]

These cases are disjoint except at addresses $t$ and $t \conc f_A$.
We have \[T(t) = E(t)\] by clause (1), and \[T(t) = A(t)\] by clause
(2).  Similarly, we have \[T(t \conc f_A) = A(f_A)\] by clause (2) and
\[T(t \conc f_A) = E(t)\] by clause (3).  So for an adjunction to be
well defined, it must be the case that \[E(t) = A(t) = A(f_A)\] that
is, the node at which adjunction occurs must have the same label as
the root and foot of the auxiliary tree adjoined.  This is, of course,
standard in definitions of TAG.

Alternatively, this constraint can be added as a stipulation and the
definition modified as follows:
\[ T(r) = \left\{ \begin{array}{ll}
			E(r) & \mbox{ if $t \not \prefixeq r$} \\
			A(u) & \mbox{ if $r = t \conc u$ and $f_A
					\not \prefixeq u$} \\
			E(t \conc u) & \mbox{ if $r = t \conc f_A
					\conc u$}
		  \end{array} \right.
\]
We will use this latter definition below.

\subsubsection{Ordered Derivation Trees}

Ordered derivation trees are ordered trees composed of nodes,
conventionally notated as $\eta$, possibly in its subscripted and
primed variants.  (For ordered derivation trees, we will be less
formal as to their mathematical structure.  In particular, the
formalization of the previous section need not apply; the definitions
that follow define all of the structure that we will need.)  The
parent of a node $\eta$ in a derivation tree will be written
$parent(\eta)$, and the tree in $\Etree$ that the node marks
adjunction of will be notated $tree(\eta)$.  The tree $tree(\eta)$ is
to be adjoined into its parent $tree(parent(\eta))$ at an address
specified on the arc in the tree linking the two; this address is
notated $addr(\eta)$.  (Of course the root node has no parent or
address; the $parent$ and $addr$ functions are partial.)

An ordered derivation tree is well-formed if for each arc in the
derivation tree from $\eta$ to $parent(\eta)$ labeled with
$addr(\eta)$, the tree $tree(\eta)$ is an auxiliary tree that can be
adjoined at the node $addr(\eta)$ in $tree(parent(\eta))$.

We repeat from Section~\ref{sec:odt} the definition of the function
$\DD$ from derivation trees to the derived trees they specify, in the
notation of this appendix:
\[
\DD(D) = \left\{
\begin{array}{l}
\makebox[0pt][l]{$tree(\eta)$}
\mbox{\qquad\qquad if $D$ is a trivial tree of one node $\eta$}\\[1ex]
tree(\eta)[\DD(D_1)/t_1, \DD(D_2)/t_2, \ldots, \DD(D_k)/t_k] \\
\mbox{\qquad\qquad if\begin{tabular}[t]{l}
		$D$ is a tree with root node $\eta$ \\
		and with $k$ child subtrees $D_1, \ldots, D_k$\\
		whose arcs are labeled with addresses $t_1, \ldots, t_k$.
		\end{tabular}}
\end{array}
\right.
\]

As in Section~\ref{sec:odt}, $E[A_1/t_1, \ldots, A_k/t_k]$ specifies
the simultaneous adjunction of trees $A_1$ through $A_k$ at $t_1$
through $t_k$, respectively, in $E$.  It is defined as the iterative
adjunction of the $A_i$ in order at their respective addresses, with
appropriate updating of the tree addresses of later adjunctions to
reflect the effect of earlier adjunctions.  In particular, the
following inductive definition suffices; the base case holds for the
adjunction of zero auxiliary trees.
\[  \begin{array}{l}
E[\,] = E \\
E[A_1/t_1, A_2/t_2, \ldots, A_k/t_k] \\
\qquad	= (E[A_1/t_1])[A_2/update(t_2, A_1, t_1),
			 \ldots, A_k/update(t_k, A_1, t_1)]
\end{array}\]

\noindent where

\[ update(s, A, t) = \left\{ \begin{array}{ll}
			s & \mbox{if $t \not \prefix s$} \\
			t \conc f_A \conc (s - t) &
				\mbox{if  $t \prefix s$}
				\end{array} \right.
\]

In the following section, we leave out parentheses in specifying
sequential adjunctions such as $(E[A_1/t_1])[A_2/t_2]$ under a
convention of left associativity of the $[\dumarg/\dumarg]$ operator.

\subsection{Effect of Sibling Swaps}

In this section, we show that the derived tree specified by a given
ordered derivation tree is unchanged if adjacent siblings whose arcs
are labeled with different tree addresses are swapped.  This will be
shown as the following proposition.
\begin{proposition}
If $t \not = t'$ then $E[ \ldots, A/t, B/t', \ldots] = E[ \ldots,
B/t', A/t, \ldots]$.
\end{proposition}

We start with a lemma, the case for only two adjunctions.
\begin{lemmanonum}
If $t \not = t'$ then $E[ A/t, B/t'] = E[ B/t', A/t]$.
\end{lemmanonum}

\proof There are three major cases, depending on the relationship of
$t$ and $t'$:
\begin{description}

\item[Case $t \prefix t'$:]  Let $s = t' - t$.  Then

\[ \begin{array}{ll}
E[A/t, B/t'](r) & = E[A/t][B/ update(t', A, t)](r) \\
	& = E[A/t][B/t \conc f_A \conc s](r) \\
	& = \left\{ \begin{array}{ll}
			E[A/t](r) & \mbox{ if $t \conc f_A \conc s
				\not \prefixeq r$} \\
			B(u) & \mbox{ if $r = t \conc f_A \conc s
					\conc u$ and $f_B
					\not \prefixeq u$} \\
			E[A/t](t \conc f_A \conc s \conc u) &
					\mbox{ if $r = t \conc f_A
					\conc s \conc f_B \conc u$}
		  \end{array} \right. \\

	& = \left\{ \begin{array}{ll}
			E(r) & \mbox{ if $t \conc f_A \conc s
				\not \prefixeq r$ and $t \not
				\prefixeq r$} \\

			A(v) & \mbox{ if $t \conc f_A \conc s
				\not \prefixeq r$ and $r = t \conc v$} \\
			E(t \conc v) & \mbox{ if $t \conc f_A \conc s
				\not \prefixeq r$ and $r = t \conc
				f_A \conc v$} \\
			B(u) & \mbox{ if $r = t \conc f_A \conc s
					\conc u$ and $f_B
					\not \prefixeq u$} \\
			E(t \conc s \conc u) &
					\mbox{ if $r = t \conc f_A
					\conc s \conc f_B \conc u$}
		  \end{array} \right. \\

	& = \left\{ \begin{array}{ll}
			E(r) & \mbox{ if $t \not \prefixeq r$} \\

			A(v) & \mbox{ if $r = t \conc v$} \\
			E(t \conc v) & \mbox{ if $s \not \prefixeq v$
					and $r = t \conc
					f_A \conc v$} \\
			B(u) & \mbox{ if $r = t \conc f_A \conc s
					\conc u$ and $f_B
					\not \prefixeq u$} \\
			E(t \conc s \conc u) &
					\mbox{ if $r = t \conc f_A
					\conc s \conc f_B \conc u$}
		  \end{array} \right.
\end{array}\]

If siblings are swapped,
\[ \begin{array}{ll}
E[B/t', A/t](r) & = E[B/t'][A/ update(t, B, t')](r) \\
	& = E[B/t'][A/t](r) \\
	& = \left\{ \begin{array}{ll}
			E[B/t \conc s](r) & \mbox{ if $t \not \prefixeq r$} \\
			A(v) & \mbox{ if $r = t \conc v$ and $f_A
					\not \prefixeq v$} \\
			E[B/t \conc s](t \conc v) &
					\mbox{ if $r = t \conc f_A
					\conc v$}
		  \end{array} \right. \\

	& = \left\{ \begin{array}{ll}
			E(r) & \mbox{ if $t \not \prefixeq r$} \\

			A(v) & \mbox{ if $r = t \conc v$} \\
			E(t \conc v) & \mbox{ if $r = t \conc
				f_A \conc v$ and $t \conc s
				\not \prefixeq t \conc v$} \\
			B(u) & \mbox{ if $r = t \conc f_A \conc v$ and
				$t \conc v = t \conc s \conc u$ and $f_B
					\not \prefixeq u$} \\
			E(t \conc s \conc u) &
					\mbox{ if $r = t \conc f_A
					\conc v$ and $t \conc v = t
					\conc f_B \conc u$ }
		  \end{array} \right. \\

	& = \left\{ \begin{array}{ll}
			E(r) & \mbox{ if $t \not \prefixeq r$} \\

			A(v) & \mbox{ if $r = t \conc v$} \\
			E(t \conc v) & \mbox{ if $s \not \prefixeq v$
					and $r = t \conc
					f_A \conc v$} \\
			B(u) & \mbox{ if $r = t \conc f_A \conc s
					\conc u$ and $f_B
					\not \prefixeq u$} \\
			E(t \conc s \conc u) &
					\mbox{ if $r = t \conc f_A
					\conc s \conc f_B \conc u$}
		  \end{array} \right.
\end{array}\]

\item[Case $t' \prefix t$:]  Analogously.

\item[Case $t \not \prefix t'$ and $t' \not \prefix t$:] \mbox{}

\[ \begin{array}{ll}
E[A/t, B/t'](r) & = E[A/t][B/ update(t', A, t)](r) \\
	& = E[A/t][B/t'](r) \\
	& = \left\{ \begin{array}{ll}
			E[A/t](r) & \mbox{ if $t'\not \prefixeq r$} \\
			B(u) & \mbox{ if $r = t' \conc u$ and $f_B
					\not \prefixeq u$} \\
			E[A/t](t' \conc u) &
					\mbox{ if $r = t' \conc f_B
					\conc u$}
		  \end{array} \right. \\
	& = \left\{ \begin{array}{ll}
			E(r) & \mbox{ if $t' \not \prefixeq r$ and $t
				\not \prefixeq r$} \\

			A(v) & \mbox{ if $t' \not \prefixeq r$ and $r
				= t \conc v$ and $f_A \not \prefixeq v$} \\
			E(t \conc v) & \mbox{ if $t' \not \prefixeq r$
					and $r = t \conc
					f_A \conc v$} \\
			B(u) & \mbox{ if $r = t' \conc u$ and $f_B
					\not \prefixeq u$} \\
			E(t' \conc u) &
					\mbox{ if $r = t' \conc f_B
					\conc u$}
		  \end{array} \right.
\end{array}\]

Note that this is unchanged (up to variable renaming) under swapping
of $A$ for $B$ and $t$ for $t'$.  That is \(E[A/t, B/t'](r) = E[B/t',
A/t](r)\).\mbox{}\qed
\end{description}

We now return to the main proposition.
\begin{proposition}
If $t \not = t'$ then $E[ \ldots, A/t, B/t', \ldots] = E[ \ldots,
B/t', A/t, \ldots]$.
\end{proposition}

\proof  The effect of the adjunctions before the two specified in the
swap is obviously the same on all following adjunctions, so we need
only show that
\[E[A/t, B/t', C_1/t_1, \ldots, C_k/t_k] = E[B/t', A/t, C_1/t_1,
\ldots, C_k/t_k ]\]
without loss of generality.  We examine the effect of the $A$ and $B$
adjunctions on the tree address $t_i$ for each $C_i$ separately.
In the case of the former adjunction order
\[\begin{array}{ll}
\multicolumn{2}{l}{E[A/t, B/t', \ldots, C_i/t_i, \ldots]} \\
\qquad & = E[A/t][B/update(t',A,t), \ldots, C_i/update(t_i,A,t), \ldots] \\
&  = E[A/t][B/update(t',A,t)][\ldots,
	C_i/update(update(t_i,A,t),B,update(t',A,t)), \ldots] \\
&  = E[A/t, B/t'][\ldots,
	C_i/update(update(t_i,A,t),B,update(t',A,t)), \ldots]
\end{array}\]
and for the latter adjunction order:
\[\begin{array}{ll}
\multicolumn{2}{l}{E[B/t', A/t, \ldots, C_i/t_i, \ldots]}\\
\qquad &  = E[B/t'][A/update(t,B,t'), \ldots, C_i/update(t_i,B,t'), \ldots] \\
&  = E[B/t'][A/update(t,B,t')][\ldots,
	C_i/update(update(t_i,B,t'),A,update(t,B,t')), \ldots] \\
&  = E[B/t', A/t][\ldots,
	C_i/update(update(t_i,B,t'),A,update(t,B,t')), \ldots] \\
&  = E[A/t, B/t'][\ldots,
	C_i/update(update(t_i,B,t'),A,update(t,B,t')), \ldots]
\end{array}\]
This last step holds by virtue of the lemma.

Thus, it suffices to show that
\[update(update(t_i,A,t),B,update(t',A,t))  \\
    = update(update(t_i,B,t'),A,update(t,B,t'))\]

Again, we perform a case analysis depending on the prefix
relationships of $t$, $t'$, and $t_i$.  Note that we make use of the
fact that if $t \prefix t'$ then $(t' - t) \conc s = t' \conc s - t$.
\begin{description}

\item[Case $t \prefix t'$:] \mbox{}

\begin{description}

\item[Subcase $t' \prefix t_i$:] \mbox{}

\[\begin{array}{l}
update(update(t_i,A,t),B,update(t',A,t)) \\
\qquad = update(t \conc f_A \conc (t_i - t), B, t \conc f_A \conc (t'
- t)) \\
\qquad = t \conc f_A \conc (t'-t) \conc f_B \conc (t_i - t') \\
\qquad = t \conc f_A \conc (t' \conc f_B \conc (t_i - t') - t) \\
\qquad = update(t' \conc f_B \conc (t_i -t'), A, t) \\
\qquad = update(update(t_i,B,t'),A,update(t,B,t'))
\end{array}\]

\item[Subcase $t' \not \prefix t_i$ and $t \prefix t_i$:] \mbox{}

\[\begin{array}{l}
update(update(t_i,A,t),B,update(t',A,t)) \\
\qquad = update(t \conc f_A \conc (t_i - t), B, t \conc f_A \conc (t'-t)) \\
\qquad = t \conc f_A \conc (t_i-t) \\
\qquad = update(t_i, A, t) \\
\qquad = update(update(t_i,B,t'),A,update(t,B,t'))
\end{array}\]

\item[Subcase $t' \not \prefix t_i$ and $t \not \prefix t_i$:] \mbox{}

\[\begin{array}{l}
update(update(t_i,A,t),B,update(t',A,t)) \\
\qquad = update(t_i, B, t \conc f_A \conc (t' - t)) \\
\qquad = t_i \\
\qquad = update(t_i, A, t \conc f_B \conc (t'-t)) \\
\qquad = update(update(t_i,B,t'),A,update(t,B,t'))
\end{array}\]

\end{description}

\item[Case $t' \prefix t$:] The proof is as for the previous
case with $t$ for $t'$ and vice versa.

\item[Case $t \not \prefix t'$ and $t' \not \prefix t$:] \mbox{}

\begin{description}

\item[Subcase $t \prefix t_i$:]  We can conclude from the assumptions
that $t' \not \prefix t_i$.  Then

\[\begin{array}{l}
update(update(t_i,A,t),B,update(t',A,t)) \\
\qquad = update(t \conc f_A \conc (t_i - t), B, t') \\
\qquad = t \conc f_A \conc (t_i - t) \\
\qquad = update(t_i, A, t) \\
\qquad = update(update(t_i,B,t'),A,update(t,B,t'))
\end{array}\]

\item[Subcase $t \not \prefix t_i$ and $t' \prefix t_i$:]  The proof
is as for the previous subcase with $t$ for $t'$ and vice
versa.

\item[Subcase $t \not \prefix t_i$ and $t' \not \prefix t_i$:] \mbox{}

\[\begin{array}{l}
update(update(t_i,A,t),B,update(t',A,t)) \\
\qquad = update(t_i, B, t') \\
\qquad = t_i \\
\qquad = update(t_i, A, t) \\
\qquad = update(update(t_i,B,t'),A,update(t,B,t'))
\end{array}\]

\end{description}\mbox{}\qed

\end{description}

\end{document}